\documentclass{article} % For LaTeX2e
\usepackage{iclr2020_conference,times}

\usepackage{amsmath,amsfonts,bm}

\newcommand{\G}{\mathcal{G}}
\newcommand{\N}{\mathcal{N}}
\newcommand{\V}{\mathcal{V}}
\newcommand{\E}{\mathcal{E}}

\newcommand{\xhdr}[1]{{\noindent\bfseries #1}.}
\newcommand{\cut}[1]{}

\newcommand{\joey}{\textcolor{blue}}
\newcommand{\mb}{\mathbf}
\newcommand{\removelatexerror}{\let\@latex@error\@gobble}

\newcommand{\GNN}{\textsc{GNN}}

\def\eqref#1{equation~\ref{#1}}
\def\1{\bm{1}}

\DeclareMathAlphabet{\mathsfit}{\encodingdefault}{\sfdefault}{m}{sl}
\SetMathAlphabet{\mathsfit}{bold}{\encodingdefault}{\sfdefault}{bx}{n}

\usepackage{hyperref}
\usepackage{url}
\usepackage{graphicx}
\usepackage{graphicx}
\usepackage{subfigure}
\usepackage{amsmath}
\usepackage{amsfonts}
\usepackage{enumitem}
\usepackage{amsthm}

\usepackage{xcolor}
\usepackage{natbib}
\usepackage{bbm}
\usepackage{booktabs} 
\usepackage[ruled,vlined]{algorithm2e}

\title{Meta-Graph: Few shot Link Prediction via Meta Learning}

\author{%
  Avishek Joey Bose \thanks{Work done during Uber AI Internship.} \\
  McGill University, Mila \\
  \texttt{joey.bose@mail.mcgill.ca} \\
  \And
  Ankit Jain \\
  Uber AI \\
 \texttt{ankit.jain@uber.com} \\
  \And
  Piero Molino \\
  Uber AI \\
  \texttt{piero.molino@uber.com} \\
  \And
  William L. Hamilton\\
  McGill University, Mila \\
  \texttt{wlh@cs.mcgill.ca} \\
}
\iclrfinalcopy % Uncomment for camera-ready version, but NOT for submission.
\begin{document}

\maketitle

\begin{abstract}
We consider the task of {\em few shot link prediction}, where the goal is to predict missing edges across multiple graphs using only a small sample of known edges. 
We show that current link prediction methods are generally ill-equipped to handle this task---as they cannot effectively transfer knowledge between graphs in a multi-graph setting and are unable to effectively learn from very sparse data. To address this challenge, we introduce a new gradient-based meta learning framework, {\em Meta-Graph}, that leverages higher-order gradients along with a learned graph signature function that conditionally generates a graph neural network initialization.  
Using a novel set of few shot link prediction benchmarks, we show that {\em Meta-Graph} enables not only fast adaptation but also better final convergence and can effectively learn using only a small sample of true edges.
\end{abstract}

\section{Introduction}
Given a graph representing known relationships between a set of nodes, the goal of link prediction is to learn from the graph and infer novel or previously unknown relationships \citep{Liben-Nowell:2003:LPP:956863.956972}. For instance, in a social network we may use link prediction to power a friendship recommendation system \citep{Aiello:2012:FPH:2180861.2180866}, or in the case of biological network data we might use link prediction to infer possible relationships between drugs, proteins, and diseases \citep{zitnik2017predicting}. 
However, despite its popularity, previous work on link prediction generally focuses only on one particular problem setting: it generally assumes that link prediction is to be performed on a single large graph and that this graph is relatively complete, i.e., that at least 50\% of the true edges are observed during training \citep[e.g., see][]{grover2016node2vec,kipf2016variational,Liben-Nowell:2003:LPP:956863.956972,lu2011link}.
%This setting is appropriate for some domains, such as friendship recommendation in a densely-known social network, but it fails to capture some of the key challenges of many real-world link prediction tasks. 

In this work, we consider the more challenging setting of {\em few shot link prediction}, where the goal is to perform link prediction on  multiple graphs that contain only a small fraction of their true, underlying edges. 
This task is inspired by applications where we have access to multiple graphs from a single domain but where each of these individual graphs contains only a small fraction of the true, underlying edges. 
For example, in the biological setting, high-throughput interactomics offers the possibility to estimate thousands of biological interaction networks from different tissues, cell types, and organisms \citep{barrios2005high}; however, these estimated relationships can be noisy and sparse, and we need learning algorithms that can leverage information across these multiple graphs in order to overcome this sparsity. 
Similarly, in the e-commerce and social network settings, link prediction can often have a large impact in cases where we must quickly make predictions on sparsely-estimated graphs, such as when a service has been recently deployed to a new locale. In other words, link prediction for a new sparse graph can benefit from transferring knowledge from other, possibly more dense, graphs assuming there is exploitable shared structure.

We term this problem of link prediction from sparsely-estimated multi-graph data as few shot link prediction analogous to the popular few shot classification setting \citep{miller2000learning,lake2011one,koch2015siamese}. 
The goal of few shot link prediction is to observe many examples of graphs from a particular domain and leverage this experience to enable fast adaptation and higher accuracy when predicting edges on a new, sparsely-estimated graph from the same domain---a task that can can also be viewed as a form of meta learning, or learning to learn \citep{bengio1990learning,bengio1992optimization,thrun2012learning,schmidhuber1987evolutionary} in the context of link prediction.
This few shot link prediction setting is particularly challenging as current link prediction methods are generally ill-equipped to transfer knowledge between graphs in a multi-graph setting and are also unable to effectively learn from very sparse data.

\xhdr{Present work} We introduce a new framework called {\em Meta-Graph} for few shot link prediction and also introduce a series of benchmarks for this task. 
We adapt the classical gradient-based meta-learning formulation for few shot classification \citep{miller2000learning,lake2011one,koch2015siamese} to the graph domain. 
Specifically, we consider a distribution over graphs as the distribution over tasks from which a global set of parameters are learnt, and we deploy this strategy to train graph neural networks (GNNs) that are capable of few-shot link prediction.
To further bootstrap fast adaptation to new graphs we also introduce a graph signature function, which learns how to map the structure of an input graph to an effective initialization point for a GNN link prediction model. 
We experimentally validate our approach on three link prediction benchmarks. 
We find that our Meta-Graph approach not only achieves fast adaptation but also converges to a better overall solution in many experimental settings, with an average improvement of $5.3 \%$ in AUC at convergence over non-meta learning baselines.

\cut{
One of the hallmarks of human intelligence is {\em fast adaptation}, i.e., the ability to learn and adapt to novel tasks when presented with minimal evidence. 
%In contrast, traditional deep learning approaches require rigid modelling of specific tasks and an abundance of data, making them inflexible
%to easily
%and hard to
%adapt to new tasks which share some similarity to the previously trained one. 
Gradient based meta-learning approaches \citep{andrychowicz2016learning,ravi2016optimization} attempt to achieve fast adaptation by learning a set of global parameters that are shared across tasks and that can then be used as a good initialization for new, related tasks.
These meta-learning approaches have achieved state-of-the-art results across numerous fast adaptation tasks in recent years, e.g., for reinforcement learning \citep{finn2017model} and few-shot image classification \citep{finn2017model,ravi2016optimization}.

However, while meta-learning algorithms have successfully been deployed in traditional deep learning domains such as images \citep{santoro2016meta}, or text \citep{madotto2019personalizing}, graph-structured data has received little attention. Meta-learning for graph structured data represents a practical setting for many real-world problems where only a limited set of sub-graphs from a larger graph are available and retraining a model is computationally expensive.
Furthermore, when only a sparse subset of edges are observable---e.g., in a new social network---meta-learning can be a viable solution for effective recommendation provided there are training graphs that share structural similarities with the new sparse one.

\xhdr{Present Work}
In this work, we adapt the classical gradient-based meta-learning formulation for few shot classification to the graph domain. Specifically, we consider a distribution over graphs as the distribution over tasks from which a global set of parameters is learned. 
We consider the challenging task of predicting missing links for each graph (i.e., link prediction) when only a small fraction of the edges can be observed. To further bootstrap fast adaptation to new graphs we also introduce a graph signature that utilizes the similarity, from the perspective of meta-learning, between the new graph and the previous graphs seen during training. We experimentally validate our approach on two standard graph datasets.
We find that our meta-learning based approach successfully achieves fast adaptation, while also converging to better overall solutions in many experimental settings. 
}
\begin{figure}[ht]
    \begin{center}
    \includegraphics[width=1\linewidth]{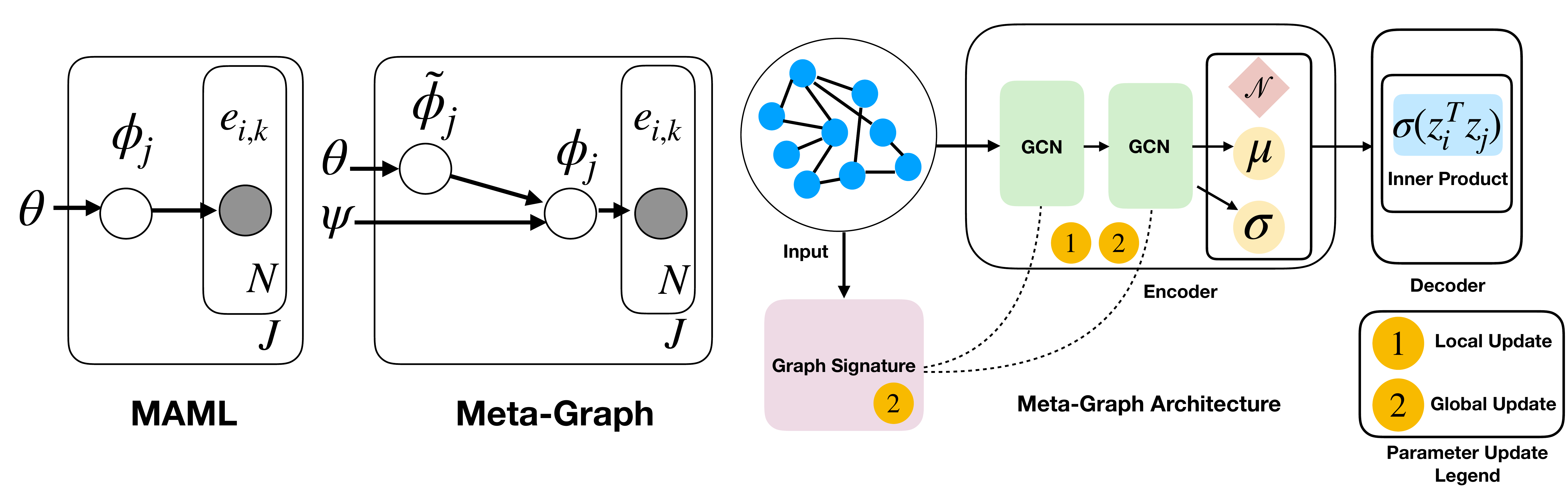}
    \end{center}
    \caption{\textbf{Left:} Graphical model for Meta-Graph vs. MAML. \textbf{Right:} Meta-Graph architecture.}
    \label{fig:meta-graph-arch}
    \vspace{-10pt}
\end{figure}
\section{Preliminaries and Problem Definition}\label{sec:background}
The basic set-up for few shot link prediction is as follows:
We assume that we have a distribution $p(\G)$ over graphs, from which we can sample training graphs $\G_i \sim p(\G)$, where each $\G_i = (\V_i, \E_i, {X}_i)$ is defined by a set of nodes $\V_i$, edges $\E_i$, and matrix of real-valued node attributes $X \in \mathbb{R}^{|\V_i| \times d}$.
When convenient, we will also equivalently represent a graph as $\G_i = (\V_i, A_i, {X}_i)$, where $A_i \in \mathbb{Z}^{|\V_i| \times |\V_i|}$ is an adjacency matrix representation of the edges in $\E_i$. We assume that each of these sampled graphs, $\G_i$, is a simple graph (i.e., contain a single type of relation and no self loops) and that every node  $v \in \V_i$ in the graph is associated with a real valued attribute vector $\mb{x}_v \in \mathbb{R}^d$ from a common vector space. We further assume that for each graph $\G_i$ we have access to only a sparse subset of the true edges $\E^{\textrm{train}}_i \subset \E_i$ (with $|\E^{\textrm{train}}_i| << |\E_i|$) during training. In terms of distributional assumptions we assume that this $p(\G)$ is defined over a set of related graphs (e.g., graphs drawn from a common domain or application setting).

%We further assume that for each of these sampled graphs, $\G_i$, we only have access to a sparse subset of the true edges $\E^{\textrm{train}}_i \subset \E_i$ (with $|\E^{\textrm{train}}_i| << |\E_i|$) during training. 
%In terms of distributional assumptions on $p(\G)$, we assume that this distribution is defined over a set of related graphs (e.g., graphs drawn from a common domain or application setting), that all the graphs are simple graphs (i.e., contain a single type of relation and no self loops), and that every node $v \in \V_i$ in each graph is associated with a real valued attribute vector $\mb{x}_v \in \mathbb{R}^d$ from a common vector space. 

Our goal is to learn a {\em global} or {\em meta} link prediction model from a set of sampled training graphs $\G_i \sim p(\G), i=1...n$, such that we can use this meta model to quickly learn an effective link prediction model on a newly sampled graph $\G_* \sim p(\G)$.
More specifically, we wish to optimize a global set of parameters $\theta$, as well as a graph signature function $\psi(\G_*)$, which can be used together to generate an effective parameter initialization, $\phi_*$, for a {\em local} link prediction model on graph $\G_*$. 
%That is, we want to optimize the global model $\psi$ such that we can produce a set of local parameters $\phi_i = \psi(\G_i)$, which can be used to initialize a local link prediction model $g_{\phi_i}$ on the graph $\G_i$, and where the local parameters $\phi$ may be fine-tuned on graph $\G_i$ using gradient descent. 

\xhdr{Relationship to standard link prediction}
Few shot link prediction differs from standard link prediction in three important ways:
\begin{enumerate}[leftmargin=*, itemsep=2pt, topsep=0pt, parsep=0pt]
    \item Rather than learning from a single graph $\G$, we are learning from multiple graphs $\{\G_1, ..., \G_n\}$ sampled from a common distribution or domain. 
    \item We presume access to only a very sparse sample of true edges. Concretely, we focus on settings where at most 30\% of the edges in $\E_i$ are observed during training, i.e., where $\frac{|\E^{\textrm{train}}|}{|\E|} \leq 0.3$.\footnote{By ``true edges'' we mean the full set of ground truth edges available in a particular dataset.}
   % Previous work on link prediction generally assumes that at least 50\% of the edges are given for training \citep.\footnote{ADD CAVEAT ON ``TRUE EDGES''}
    \item We distinguish between the {\em global} parameters, which are used to encode knowledge about the underlying distribution of graphs, and the {\em local} parameters $\phi_i$, which are optimized to perform link prediction on a specific graph $\G_i$.
    This distinction allows us to consider leveraging information from multiple graphs, while still allowing for individually-tuned link prediction models on each specific graph. 
\end{enumerate}

\xhdr{Relationship to traditional meta learning}
Traditional meta learning for few-shot classification generally assumes a distribution $p(\mathcal{T})$ over classification tasks, with the goal of learning global parameters that can facilitate fast adaptation to a newly sampled task $\mathcal{T}_i \sim p(\mathcal{T})$ with few examples. %so that a model can quickly classify examples for a newly sampled task $\mathcal{T}_i \sim p(\mathcal{T})$ \citep. 
We instead consider a distribution $p(\G)$ over graphs with the goal of performing link prediction on a newly sampled graph.
An important complication of this graph setting is that the individual predictions for each graph (i.e., the training edges) are not i.i.d.. Furthermore, for few shot link prediction we require training samples as a sparse subset of true edges that represents a small percentage of all edges in a graph. Note that for very small percentages of training edges we effectively break all graph structure and recover the supervised setting for few shot classification.% and thus simplifying the problem. 

\cut{
The key idea behind MetaGraph is to augment a GNN link prediction model using gradient-based meta learning, as well as with a novel graph signature function that learns to conditionally generate parameter initializations based upon graph structure.  

\cut{
\xhdr{Link Prediction}
Link prediction (LP) is a problem of predicting whether two nodes are likely to have an edge in the graph \citep{Liben-Nowell:2003:LPP:956863.956972}. LP has been used in many applications like friend and content recommendations \citep{Aiello:2012:FPH:2180861.2180866} , shopping and movie recommendation \citep{Huang:2005:LPA:1065385.1065415} , knowledge graph completion \citep{nickel2015review} and even to identify criminals based on their activities \citep{Hasan06linkprediction}. Historically, link prediction has been mainly performed using topological features of the graph like common neighbors. This has yielded very strong baselines like Adamic/Adar measure \citep{adamic2003friends}, Jaccard Index among others. Other approaches include Matrix Factorization \citep{Menon:2011:LPV:2034117.2034146} and more recently graph neural networks \citep{zhang2018link} have been also been explored for this application. However, all of the above mentioned approaches dealt with a single dense graph where you try to predict unknown/future links. Unlike these previous approaches, our approach considers link prediction tasks on multiple sparse graphs which are drawn from similar real world scenario. 
}

\xhdr{Variational Graph Autoencoder}
One of the most prominent approaches to unsupervised learning is the Variational Autoencoder (VAE) \citep{kingma2013auto}.
%and analagously the Variational Graph Autoencoder (VGAE) if the data is graph structured.
For data structured as a graph, an analogous Variational Graph Autoencoder (VGAE) has been proposed \citep{kipf2016variational}.
Formally, given a graph $\mathcal{G}=(V,\E) $, with $N = |V|$ nodes, a weight matrix $W$, an adjacency matrix $A$ and node feature matrix $X \in \mathbb{R}^{N \times D}$, the VGAE learns both an inference model that effectively encodes each node into an embedding vector as well as a generative model that scores the likelihood of an edge existing between pairs of nodes. The parameters of the inference or recognition network are shared across all nodes in $\mathcal{G}$, effectively \textit{amortizing} the inference process needed to define the approximate posterior, $q(z|X,A)= \prod_{i=1}^N q(z_i|X,A)$ where $q(z|X,A)=\mathcal{N}(z_i|\mu_i,\textnormal{diag}(\sigma_i^2))$. Conversely, the generative network models is defined by $p(A|Z) = \prod_{i=1}^N \prod_{j=1}^N p(A_{i,j}|z_i,z_j)$, that is the likelihood of an edge existing between the node pairs. In much the same vein, the overall loss function optimizes for the variational lower bound given by:
\begin{equation}\label{eq:vgae_loss}
\centering
\mathcal{L}_{G} = \mathbb{E}_{q}[\log p(A|Z)] - KL[q(Z|X,A) || p(z)]
\end{equation}
Optimizing, this lower bound effectively maximizes the log likelihood of the data as the $KL-$divergence is positive by definition.

\xhdr{Meta-Learning}
Humans have a remarkable ability to perform new tasks having only been briefly exposed to them. Much of this is attributed to the fact that skills that were learned in previous experiences can be reused, and thus not relearned from scratch, to bootstrap the learning process in a new task. In meta-learning or learning to learn \citep{bengio1990learning,bengio1992optimization,thrun2012learning,schmidhuber1987evolutionary}, the objective is to learn from prior experiences to form inductive biases for fast adaptation to unseen tasks. While there are many approaches to meta-learning, in this work we focus on a class of approaches known as gradient-based meta-learning, where stochastic gradient descent is used to backpropagate through the learning process itself. Let $\mathcal{D}$ be a dataset which defines a distribution over tasks $\mathcal{T}$ with some shared structure and, each task $\mathcal{T}_j$ defines a distribution over datapoints, $x_j$, which is to be optimized by the meta-learner to produce task specific parameters. In the few-shot learning setting, the meta-learner observes up to $N$ samples from each task ---i.e. $x_{j_1}, ... , x_{j_N} \sim p(\mathcal{T}_{j})$ with the goal of finding a set of shared parameters, $\theta$, over tasks. These global parameters are optimized such that when a few gradient descent steps are taken from the initialization, $\theta$, given a small sample of points from $\mathcal{T}_{j}$ there is good generalization performance on held out samples also from the same task---i.e., $x_{j_{N+1}}, ... , x_{j_{N+m}} \sim p(\mathcal{T}_{j_i})$. That is, starting from the global parameters $\theta$ the meta-learning algorithm produces a set of local parameters, $\phi_j$ tailored to $\mathcal{T}_j$, through fast adaptation. This approach to gradient based meta-learning is known as Model Agonistic Meta-Learning (MAML) \citep{finn2017model}, and it's overall objective is defined as:
\begin{align}
    \mathcal{L}(\theta) &= \frac{1}{J} \sum_{j \in J} \Big[\sum_{m \in M} -\log p(x_{j_{N+m}}|\theta - \alpha \nabla_{\theta} \frac{1}{N} \sum_{n \in N} - \log p(x_{j_n}|\theta)) \Big].
\end{align}
Here, the local parameters are $\phi_j = \theta - \alpha \nabla_{\theta} \frac{1}{N} \sum_{n \in N} - \log p(x_{j_n}|\theta)$ with $\alpha$ as the learning rate for the specific task.

}
\section{Proposed Approach}

We now outline our proposed approach, Meta-Graph, to the few shot link prediction problem.
We first describe how we define the local link prediction models, which are used to perform link prediction on each specific graph $\G_i$. Next, we discuss our novel gradient-based meta learning approach to define a global model that can learn from multiple graphs to generate effective parameter initializations for the local models. 
The key idea behind Meta-Graph is that we use gradient-based meta learning to optimize a shared parameter initialization $\theta$ for the local models, while also learning a parametric encoding of each graph $\G_i$ that can be used to modulate this parameter initialization in a graph-specific way (Figure \ref{fig:meta-graph-arch}).

\subsection{Local Link Prediction Model}
%Our Meta-Graph framework assumes that graph neural networks (GNNs) are used to define the local link prediction models on each graph $\G_i$. 
In principle, our framework can be combined with a wide variety of GNN-based link prediction approaches, but here we focus on variational graph autoencoders (VGAEs) \citep{kipf2016variational} as our base link prediction framework. 
Formally, given a  graph $\mathcal{G}=(\V,A, X)$, the VGAE learns an inference model, $q_\phi$, that defines a distribution over node embeddings $q_\phi(Z | A, X)$,
where each row $z_v \in \mathbb{R}^d$ of $Z \in \mathbb{R}^{|\V| \times d}$ is a node embedding that can be used to score the likelihood of an edge existing between pairs of nodes. 
The parameters of the inference model are shared across all the nodes in $\mathcal{G}$, to define the approximate posterior $q_\phi(z_v|A,X)=\mathcal{N}(z_v|\mu_v,\textnormal{diag}(\sigma_v^2))$, where  the parameters of the normal distribution are learned via GNNs:
\begin{equation}\label{eq:inferencegnns}
\mu = \GNN_\mu(A, X) , \qquad \textrm{and} \qquad  \log(\sigma) = \GNN_{\sigma}(A,X).
\end{equation}
The generative component of the VGAE is then defined as
\begin{equation}
p(A|Z) = \prod_{i=1}^N \prod_{j=1}^N p(A_{u,v}|z_u,z_v), \qquad \textrm{with} \qquad  p(A_{u,v}|z_u,z_v) = \sigma(z_u^\top z_v),
\end{equation}
i.e., the likelihood of an edge existing between two nodes, $u$ and $v$, is proportional to the dot product of their node embeddings.
Given the above components, the inference GNNs can  be trained to minimize the variational lower bound on the training data:
\begin{equation}\label{eq:vgae_loss}
\centering
\mathcal{L}_{G} = \mathbb{E}_{q_\phi}[\log p(A^{\textrm{train}}|Z)] - KL[q_\phi(Z|X,A^{\textrm{train}}) || p(z)],
\end{equation}
where a Gaussian prior is used for $p(z)$.

We build upon VGAEs due to their strong performance on standard link prediction benchmarks \citep{kipf2016variational}, as well as the fact that they have a well-defined probabilistic interpretation that generalizes many embedding-based approaches to link prediction (e.g., node2vec \citep{grover2016node2vec}). 
We describe the specific GNN implementations we deploy for the inference model in Section \ref{sec:variants}.

\subsection{Overview of Meta-Graph}
The key idea behind Meta-Graph is that we use gradient-based meta learning to optimize a shared parameter initialization $\theta$ for the inference models of a VGAE, while also learning a parametric encoding $\psi(\G_i)$ that modulates this parameter initialization in a graph-specific way.
Specifically, given a sampled training graph $\G_i$, we initialize the inference model $q_{\phi_i}$ for a VGAE link prediction model using a combination of two learned components:
\begin{itemize}[leftmargin=*, itemsep=2pt, topsep=0pt, parsep=0pt]
  \item A global initialization, $\theta$, that is used to initialize all the parameters of the GNNs in the inference model. The global parameters $\theta$ are optimized via second-order gradient descent to provide an effective initialization point for any graph sampled from the distribution $p(\G)$.
  \item A graph signature $s_{\G_i} = \psi(\G_i)$ that is used to modulate the parameters of inference model $\phi_{i}$ based on the history of observed training graphs.\cut{for graph $\G_i$ in a graph-specific way.} In particular, we assume that the inference model $q_{\phi_i}$ for each graph $\G_i$ can be conditioned on the graph signature. That is, we augment the inference model to $q_{\phi_i}(Z |A, X, s_{\G_i})$, where we also include the graph signature $s_{\G_i}$ as a conditioning input. 
  We use a k-layer graph convolutional network (GCN) \citep{kipf2016semi}, with sum pooling to compute the signature:
  \begin{equation}\label{eq:sig}
      s_\G = \psi(\G) = \textrm{MLP}(\sum_{v \in \V}z_v) \qquad \textrm{with} \qquad  Z = \textrm{GCN}(A,X),
  \end{equation}
  where $\textrm{GCN}$ denotes a k-layer GCN (as defined in \citep{kipf2016semi}), MLP denotes a densely-connected neural network, and we are summing over the node embeddings $z_v$ output from the GCN. 
  As with the global parameters $\theta$, the graph signature model $\psi$ is optimized via second-order gradient descent. 
\end{itemize}

The overall Meta-Graph architecture is detailed in Figure \ref{fig:meta-graph-arch}  and the core learning algorithm is summarized in the algorithm block below.

\begin{algorithm}[H]
\SetAlgoLined
\KwResult{Global parameters $\theta$, Graph signature function $\psi$}
 Initialize learning rates: $\alpha,\epsilon$ \\
 Sample a mini-batch of graphs, $\mathcal{G}_{batch}$ from $p(\mathcal{G})$\;
 \For{each $\mathcal{G}$ $\in \mathcal{G}_{batch}$}{
  $\mathcal{E} = \mathcal{E}^{\textrm{train}} \cup \mathcal{E}^{\textrm{val}} \cup \mathcal{E}^{\textrm{test}}$  // Split edges into train, val, and test\\ 
   $s_\G = \psi(\mathcal{G}, \mathcal{E}^{\textrm{train}})$ // Compute graph signature\\
  Initialize: $\phi^{(0)} \leftarrow \theta$ // Initialize local parameters via global parameters\\ 
   \For{$k \ in \ [1:K]$}{
       $s_\G =\textnormal{stopgrad}(s_\G)$ // Stop Gradients to Graph Signature
       $\mathcal{L}_{train} = \mathbb{E}_{q}[\log p(A^{\textrm{train}}|Z)] - KL[q_\phi(Z| \mathcal{E}^{\textrm{train}}, s_\G) || p(z)]$ \\
       Update $\phi^{(k)} \leftarrow \phi^{(k-1)} - \alpha \nabla_{\phi}\mathcal{L}_{train}$ \\ 
 }
Initialize: $\theta \leftarrow \phi_K$\\
$s_\G = \psi(\mathcal{G}, \mathcal{E}^{\textrm{val}} \cup \mathcal{E}^{\textrm{train}})$ // Compute graph signature with validation edges\\
$\mathcal{L}_{val} = \mathbb{E}_{q}[\log p(A^{\textrm{val}}|Z)] - KL[q(Z| \mathcal{E}^{\textrm{val}} \cup \mathcal{E}^{\textrm{train}} , s_\G) || p(z)]$ \\
Update $\theta \leftarrow \theta - \epsilon \nabla_{\theta}\mathcal{L}_{val}$ \\
Update $\psi \leftarrow \psi - \epsilon \nabla_{\psi}\mathcal{L}_{val}$ \\
}
\caption{Meta-Graph for Few Shot Link Prediction}
\label{alg:metagraph}
\end{algorithm}
The basic idea behind the algorithm is that we (i) sample a batch of training graphs, (ii) initialize VGAE link prediction models for these training graphs using our global parameters and signature function, (iii) run $K$ steps of gradient descent to optimize each of these VGAE models, and (iv) use second order gradient descent to update the global parameters and signature function based on a held-out validation set of edges. As depicted in Fig \ref{fig:meta-graph-arch}, this corresponds to updating the GCN based encoder for the local link prediction parameters $\phi_j$ and global parameters $\theta$ along with the graph signature function $\psi$ using second order gradients.
Note that since we are running $K$ steps of gradient descent within the inner loop of Algorithm \ref{alg:metagraph}, we are also ``meta'' optimizing for fast adaptation, as $\theta$ and $\psi$ are being trained via second-order gradient descent to optimize the local model performance after $K$ gradient updates, where generally $K \in \{0,1, \ldots ,5\}$.

\subsection{Variants of Meta-Graph}\label{sec:variants}

We consider several concrete instantiations of the Meta-Graph framework, which differ in terms of how the output of the graph signature function is used to modulate the parameters of the VGAE inference models.
For all the Meta-Graph variants, we build upon the standard GCN propagation rule \citep{kipf2016semi} to construct the VGAE inference models. 
In particular, we assume that all the inference GNNs (Equation \ref{eq:inferencegnns})  are defined by stacking $K$ neural message passing layers of the form:
\begin{equation}\label{eq:message}
    h_v^{(k)} = \textrm{ReLU}\left(\sum_{u \in \N(v) \cup \{v\}}\frac{m_{s_\G}\left(W^{(k)}h^{(k-1)}_u\right)}{\sqrt{|\N(v)||\N(u)|}}\right),
\end{equation}
where $h_v \in \mathbb{R}^d$ denotes the embedding of node $v$ at layer $k$ of the model, $\N(v) = \{u \in \V : e_{u,v} \in \E\}$ denotes the nodes in the graph neighborhood of $v$, and $W^{(k)} \in \mathbb{R}^{d \times d}$ is a trainable weight matrix for layer $k$.
The key difference between Equation \ref{eq:message} and the standard GCN propagation rule is that we add the modulation function $m_{s_\G}$, which is used to modulate the message passing based on the graph signature $s_\G = \psi(\G)$. 

We describe different variations of this modulation below. 
In all cases, the intuition behind this modulation is that we want to compute a structural signature from the input graphs that can be used to condition the initialization of the local link prediction models. 
Intuitively, we expect this graph signature to encode structural properties of sampled graphs $\G_i \sim p(\G)$ in order to modulate the parameters of the local VGAE link prediction models and adapt it to the current graph.

\xhdr{GS-Modulation}
Inspired by \citet{brockschmidt2019gnn}, we experiment with basic feature-wise linear modulation \citep{strub2018visual} to define the modulation function $m_{s_\G}$:
\begin{align}\label{eq:film}
    \beta_k,\gamma_k, &= \psi(\mathcal{G})\nonumber \\ 
    m_{\beta_k,\gamma_k}\left(W^{(k)}h^{(k-1)}_u\right) &=  \gamma_k \odot W h^{(k-1)} + \beta_k. 
\end{align}
Here, we restrict the modulation terms $\beta_k$ and $\gamma_k$ output by the signature function to be in $[-1,1]$ by applying a $\tanh$ non-linearity after Equation \ref{eq:sig}.

\xhdr{GS-Gating}
Feature-wise linear modulation of the GCN parameters (Equation \ref{eq:film}) is an intuitive and simple choice that provides flexible modulation while still being relatively constrained. 
However, one drawback of the basic linear modulation is that it is ``always on'', and there may be instances where the modulation could actually be counter-productive to learning. 
To allow the model to adaptively learn when to apply modulation, we extend the feature-wise linear modulation using a sigmoid gating term, $\rho_k$ (with $[0,1]$ entries), that gates in the influence of $\gamma$ and $\beta$:
\begin{align*}
    \beta_k,\gamma_k, \rho_k &= \psi(\mathcal{G})\\
    \beta_k & = \rho_k \odot \beta_k + (\mathbbm{1} - \rho_k) \odot \mathbbm{1} \\
    \gamma_k & = \rho_k \odot \gamma_k + (\mathbbm{1} - \rho_k) \odot \mathbbm{1} \\
      m_{\beta_k,\gamma_k}\left(W^{(k)}h^{(k-1)}_u\right) &=  \gamma_k \odot W h^{(k-1)} + \beta_k.
\end{align*}

\xhdr{GS-Weights}
In the final variant of Meta-Graph, we extend the gating and modulation idea by separately aggregating graph neighborhood information with and without modulation and then merging these two signals via a convex combination:
\begin{align*}
    \beta_k,\gamma_k, \rho_k &= \psi(\mathcal{G})\\
    h_v^{(k),1} &=  \textrm{ReLU}\left(\sum_{u \in \N(v) \cup \{v\}}\frac{W^{(k)}h^{(k-1)}_u}{\sqrt{|\N(v)||\N(u)|}}\right) \\
    h_v^{(k),2} &=  \textrm{ReLU}\left(\sum_{u \in \N(v) \cup \{v\}}\frac{m_{s_{ \beta_k,\gamma_k}}\left(W^{(k)}h^{(k-1)}_u\right)}{\sqrt{|\N(v)||\N(u)|}}\right)\\
    h^{(k)}_v & = \rho_k \odot h_v^{(k),1} + (\mathbbm{1} - \rho_k) \odot   h_v^{(k),2},
\end{align*}
where we use the basic linear modulation (Equation \ref{eq:film}) to define $m_{s_{ \beta_k,\gamma_k}}$.

\subsection{MAML for link prediction as a special case}

Note that a simplification of Meta-Graph, where the graph signature function is removed, can be viewed as an adaptation of model agnostic meta learning (MAML)  \citep{finn2017model} to the few shot link prediction setting.
As discussed in Section \ref{sec:background}, there are important differences in the set-up for few shot link prediction, compared to traditional few shot classification. 
Nonetheless, the core idea of leveraging an inner and outer loop of training in Algorithm \ref{alg:metagraph}---as well as using second order gradients to optimize the global parameters---can be viewed as an adaptation of MAML to the graph setting, and we provide comparisons to this simplified MAML approach in the experiments below. We formalize the key differences by depicting the graphical model of MAML as first depicted in \citep{grant2018recasting} and contrasting it with the graphical model for Meta-Graph, in Figure~ \ref{fig:meta-graph-arch}. MAML when reinterpreted for a distribution over graphs, maximizes the likelihood over all edges in the distribution. On the other hand, Meta-Graph when recast in a hierarchical Bayesian framework adds a graph signature function that influences $\tilde{\phi_j}$ to produce the modulated parameters $\phi_j$ from $N$ sampled edges. This explicit influence of $\psi$ is captured by the term $p(\tilde{\phi_j}|\psi,\phi_j)$ in Equation  \ref{metagraph_graph_model_eqn} below:
\begin{equation}
    p(\mathcal{E}|\theta) = \prod_j^J \left(\int \int p(\mathcal{E}_{j}| \phi_j)p(\phi_j|\psi,\tilde{\phi_j})p(\tilde{\phi_j}|\theta)d\phi_j d\tilde{\phi_j}\right)
     \label{metagraph_graph_model_eqn}
\end{equation}
For computational tractability we take the likelihood of the modulated parameters as a point estimate ---i.e., $p(\phi_j|\psi,\tilde{\phi_j}) = \delta(\psi \cdot \tilde{\phi_j})$. 
%As we show in the next section even with such a mild assumption Meta-Graph is able to outperform MAML and other approaches in many few-shot link prediction settings.

\cut{
parameters, $\theta, \phi$, that are based on the VGAE model comprising of $k$ hidden GCN \citep{kipf2016semi} layers as the recognition network and a dot-product decoder as the generative model. To exploit the shared structural and node feature similarities between graphs, we also define a Graph Signature (GS) function, $\psi$, which is used to directly influence the parameters of $\theta$ and $\phi$. In it's most basic form $\psi$ learns a graph embedding $\gamma$ and a bias $\beta$ for each GCN layer in the recognition network given a sampled graph, $\mathcal{G}_i\sim p(\mathcal{G})$. Intuitively, the role of $\psi$ is to encode the structural properties relevant for meta-learning across the distribution over graphs such that given a new $\mathcal{G}_i \sim p(\mathcal{G})$ it can thus inform the weight updates for even faster adaption. Similar to the recognition network,  we define $\psi$ using $k$-GCN layers followed by a small MLP with a non-linear activation such that the output is bounded ---i.e. $\gamma,\beta \in [-1,1]$. Inspired by Brockschmidt (2019) \citep{brockschmidt2019gnn}, we use feature-wise linear modulation \citep{strub2018visual} for each GCN layer in the recognition network as follows:
\begin{align*}
    \beta_k,\gamma_k, &= \psi(\mathcal{G})\\
    h_k &= \sum_{i \to j \in \E}( \gamma_k \odot W h_{k-1} + \beta_k).
\end{align*}
To enforce the GS to learn useful modulating parameters $\gamma$ and $\beta$ over the entire distribution, we update $\psi$, only in the outer loop. During meta-training, we use $\psi$ as a deterministic encoding for $\mathcal{G}$, and update local parameters $\phi$ with a few steps of gradient descent, while ensuring that gradients from the inner-loop updates are never used to update $\psi$ itself. Defining $\psi$ in the outer loop allows us to compute parameters that influence the parameters of the recognition prior to a single gradient step, enabling faster adaptation than vanilla meta-learning which does not explicitly leverage structural and other features of previously observed graphs. Fig. 1 Left shows the architecture of Meta-Graph while Fig. 1 Right gives the exact algorithm used to update global and local parameters, $\theta,\phi$ and the graph signature function.}

\cut{
\subsection{Meta-Graph Variants}
We introduce four additional variants to Meta-Graph where the main difference is the encoding functions to obtain $h_k$.
The intuition behind these variants is that in certain circumstances, particularly when a bigger sample from a graph is available, the structural information about the graph provided by the signature function may not be needed and may influence learning negatively.
For this reason a gating mechanism that controls the degree of influence of the signature function is introduced.

\paragraph{Signature Gating:} in this variant $\beta_k$ and $\gamma_k$ are interpolated with a vector of 1s through $\alpha$, the sigmoid of a learned parameter vector $w_c$:
\begin{align*}
	\alpha &= \sigma(w_c)\\
    \beta_k,\gamma_k &= \alpha \psi(\mathcal{G}) + (1-\alpha)1\\
    h_k &= \sum_{i \to j \in \E}( \gamma_k \odot W h_{k-1} + \beta_k).
\end{align*}

\paragraph{Weights Gating:} in this variant two sets of weights $W_a$ and $W_b$ are learned separately and used to obtain $h_k^a$ and $h_k^b$, the former using the signature function and the latter without the signature function. Both vectors are then interpolated through $\alpha$, the sigmoid of a learned parameter vector $w_c$:
\begin{align*}
	\alpha &= \sigma(w_c)\\
    \beta_k,\gamma_k &= \psi(\mathcal{G})\\
    h_k^a &= \sum_{i \to j \in \E}( \gamma_k \odot W_a h_{k-1} + \beta_k)\\
    h_k^b &= \sum_{i \to j \in \E}( W_b h_{k-1} )\\
	h_k &= \alpha h_k^a + (1-\alpha) h_k^b.
\end{align*}
}

\cut{
\xhdr{Connection to MAML}
The Meta-Graph algorithm has important differences from MAML and other conventional meta-learning domains. Meta-Graph assumes a distribution over graphs rather than specific tasks. Secondly, examples in our dataset are edges in a sparse graph which can be non-i.i.d. unlike in supervised classification or regression. The most significant difference, is however the addition of the Graph Signature function and its explicit role in influencing the recognition networks parameters during meta-training. To be more precise, new test graphs which are similar to training graphs, for the purposes of meta-learning, start at a significantly better initialization point due to feature-wise modulation of the recognition network parameters. We further formalize this intuition by depicting the graphical model of MAML as first depicted in \citep{grant2018recasting} and contrasting it with the graphical model for Meta-Graph in Fig. \ref{fig:graphical_model}. MAML reinterpreted for a distribution over graphs maximizes the likelihood over all edges in the distribution as written below in Eqn \ref{maml_graph_model_eqn}.

\begin{figure}
    \begin{minipage}{0.49\linewidth}
    \centering
    \includegraphics[width=1\linewidth]{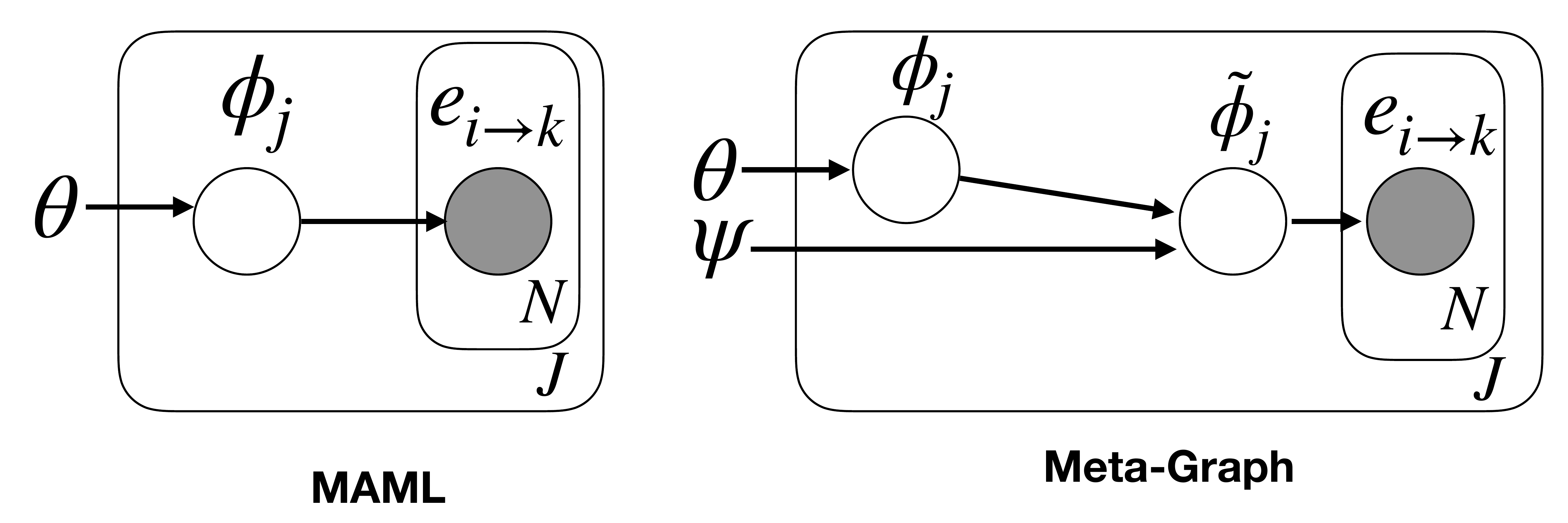}
    \caption{MAML graphical model versus Meta-Graph Graphical Model}
    \label{fig:graphical_model}
    \end{minipage}
\end{figure}

\begin{equation}
    p(\mathcal{E}|\theta) = \prod_j (\int p(\mathcal{E}_{j}| \phi_j)p(\phi_j|\theta)d\phi_j)
    \label{maml_graph_model_eqn}
\end{equation}
On the other hand, Meta-Graph when recast as hierarchical bayes adds a graph signature function that influences the local parameters of the recognition network $\phi_j$ to produce the modulated parameters $\tilde{\phi}_j$. This explicit influence of $\psi$ is captured by the term $p(\tilde{\phi_j}|\psi,\phi_j)$ in Eqn \ref{metagraph_graph_model_eqn}. 
\begin{equation}
    p(\mathcal{E}|\theta) = \prod_j^J \left(\int \int p(\mathcal{E}_{j}| \phi_j)p(\phi_j|\psi,\tilde{\phi_j})p(\tilde{\phi_j}|\theta)d\phi_j d\tilde{\phi_j}\right)
     \label{metagraph_graph_model_eqn}
\end{equation}
For computational tractability we take the likelihood of the modulated parameters as a point estimate, ---i.e. $p(\tilde{\phi_j}|\psi,\phi_j) = \delta(\psi \cdot \phi_j)$. As we show in the next section even with such a simplifying assumption Meta-Graph is able to perform better than MAML and other approaches in many few-shot link prediction settings.

}
\cut{
\begin{algorithm}[H]
\SetAlgoLined
\KwResult{Global Parameters $\theta$, Graph Signature $\psi$}
 Initialize learning rates: $\alpha,\epsilon$ \\
 Sample a mini-batch of Graphs, $\mathcal{G}_{batch}$ from $p(\mathcal{G})$\;
 \For{each Graph $\mathcal{G}$ \in \mathcal{G}_{batch}}{
  $\mathcal{E} = \mathcal{E}_{\mathcal{G}_{train}} \cup \mathcal{E}_{\mathcal{G}_{val}} \cup \mathcal{E}_{\mathcal{G}_{test}}$  // Split Edges in to train, val, and test\\ 
  Initialize: $\phi_0 \leftarrow \theta$ \;
   \For{$k \ in \ 1 , ... , K$}{
    $\beta,\gamma &= \textnormal{stopgrad}(\psi(\mathcal{G}, \mathcal{E}_{\mathcal{G}_{train}}))$ // Compute Graph Signature\\
       $\mathcal{L}_{train} = \mathbb{E}_{q}[\log p(A|Z)] - KL[q(Z| \mathcal{E}_{\mathcal{G}_i_{train}}, \beta, \gamma) || p(z)]$ \\
       Update $\phi$_k \leftarrow \phi_{k-1} - \alpha \nabla_{\phi}\mathcal{L}_{train} \\ 
 }
Initialize: $\theta \leftarrow \phi_K$\\
$\beta,\gamma &= \psi(\mathcal{G}, \mathcal{E}_{\mathcal{G}_{val}})$ // Compute Graph Signature with val edges\\
$\mathcal{L}_{val} = \mathbb{E}_{q}[\log p(A|Z)] - KL[q(Z| \mathcal{E}_{\mathcal{G}_{val}},\beta,\gamma) || p(z)]$ \\
Update $\theta$ \leftarrow \theta - \epsilon \nabla_{\theta}\mathcal{L}_{val} \\
Update $\psi$ \leftarrow \psi - \epsilon \nabla_{\psi}\mathcal{L}_{val} \\
}
\caption{Meta-Graph for Few Shot Link Prediction}
\end{algorithm}
}
\section{Experiments}
We design three novel benchmarks for the few-shot link prediction task.
All of these benchmarks contain a set of graphs drawn from a common domain.
In all settings, we use 80\% of these graphs for training and 10\% as validation graphs, where these training and validation graphs are used to optimize the global model parameters (for Meta-Graph) or pre-train weights (for various baseline approaches).
We then provide the remaining 10\% of the graphs as test graphs, and our goal is to fine-tune or train a model on these test graphs to achieve high link prediction accuracy. 
Note that in this few shot link prediction setting, {\em there are train/val/test splits at both the level of graphs and edges}: for every individual graph, we are optimizing a model using the training edges to predict the likelihood of the test edges, but we are also training on multiple graphs with the goal of facilitating fast adaptation to new graphs via the global model parameters. 

Our goal is to use our benchmarks to investigate four key empirical questions:
\begin{itemize}[leftmargin=18pt, topsep=0pt, parsep=0pt, itemsep=2pt]
    \item[{\bf Q1}] How does the overall performance of Meta-Graph compare to various baselines, including (i) a simple adaptation of MAML \citep{finn2017model} (i.e., an ablation of Meta-Graph where the graph signature function is removed), (ii), standard pre-training approaches where we pre-train the VGAE model on the training graphs before fine-tuning on the test graphs, and  (iii) naive baselines that do not leverage multi-graph information (i.e., a basic VGAE without pre-training, the Adamic-Adar heuristic \citep{adamic2003friends}, and DeepWalk \citep{perozzi2014deepwalk})? 
   \item[{\bf Q2}] How well does Meta-Graph perform in terms of fast adaption? Is Meta-Graph able to achieve strong performance after only a small number of gradient steps on the test graphs?
   \item[{\bf Q3}] How necessary is the graph signature function for strong performance, and how do the different variants of the Meta-Graph signature function compare across the various benchmark settings?
   \item[{\bf Q4}]What is learned by the graph signature function? For example, do the learned graph signatures correlate with the structural properties of the input graphs, or are they more sensitive to node feature information?
\end{itemize}

\begin{table*}[t]
\caption{Statistics for the three datasets used to test Meta-Graph.}
\begin{center}
\begin{small}
\begin{sc}
\begin{tabular}{lcccccr}
\toprule
Dataset & \#Graphs & Avg. Nodes & Avg. Edges & \#Node Feats &\\
\midrule
PPI    & 24 & 2,331 & 64,596 & 50&\\
FirstMM DB & 41& 1,377& 6,147& 5&\\
Ego-AMINER   & 72 & 462& 2245& 300&\\
\bottomrule
\end{tabular}
\end{sc}
\end{small}
\end{center}
\label{datasetstats-table}
\vskip -0.1in
\end{table*}

\xhdr{Datasets}
Two of our benchmarks are derived from standard multi-graph datasets from protein-protein interaction (PPI) networks \citep{zitnik2017predicting} and 3D point cloud data (FirstMM-DB) \citep{neumann2013graph}.
These benchmarks are traditionally used for node and graph classification, respectively, but we adapt them for link prediction.
We also create a novel multi-graph dataset based upon the AMINER citation data \citep{tang2008arnetminer}, where each node corresponds to a paper and links represent citations.
We construct individual graphs from AMINER data by sampling ego networks around nodes and create node features using embeddings of the paper abstracts (see Appendix for details). 
We preprocess all graphs in each domain such that each graph contains a minimum of $100$ nodes and up to a maximum of $20000$ nodes. %For meta-training we take $80\%$ of graphs as the training set and the remaining graphs we split evenly between validation and test.
For all datasets, we perform link prediction by training on a small subset (i.e., a percentage) of the edges and then attempting to predict the unseen edges (with $20\%$ of the held-out edges used for validation).
Key dataset statistics are summarized in Table \ref{datasetstats-table}.

\xhdr{Baseline details}
Several baselines correspond to modifications or ablations of Meta-Graph, including the straightforward adaptation of MAML (which we term {\em MAML} in the results), a finetune baseline where we pre-train a VGAE on the training graphs observed in a sequential order and fine-tune on the test graphs (termed {\em Finetune}). We also consider a VGAE trained individually on each test graph (termed {\em No Finetune}). 
For Meta-Graph and all of these baselines we employ Bayesian optimization with Thompson sampling \citep{kandasamy2018parallelised} to perform hyperparameter selection using the validation sets. 
We use the recommended default hyperparameters for DeepWalk and  Adamic-Adar baseline is hyperparameter-free. \footnote{Code is included with our submission and will be made public after the review process}

\cut{
\joey{Maybe move this section appendix to save space?}
We demonstrate the ability of Meta-Graph for effective few-shot link-prediction on the Protein Protein Interaction (PPI) \citep{zitnik2017predicting}, FirstMM DB \citep{neumann2013graph} and AMINER citation \citep{tang2008arnetminer} datasets taken from the biological, robotics and citation network domains respectively. Table \ref{datasetstats-table} contains dataset statistics such as the number of graphs, average nodes and edges per graph and the dimension of the node features. We preprocess all graphs in each domain such that each graph contains a minimum of $100$ nodes and up to a maximum of $20000$ nodes. For meta-training we take $80\%$ of graphs as the training set and the remaining graphs we split evenly between validation and test. The PPI datasets consists of human protein-protein interaction networks corresponding to different tissues. The FirstMM DB contains a set of graphs corresponding to 3d point cloud data and categories of various household objects for semantic and graph-based object category prediction. To create citation networks we first separate the data based on the field of study from which we construct each graph. We then preprocess the data by selecting the top $100$ graphs based on number of edges and extract a 5-core subgraph for each of the graphs. Finally, we pick a random node and compute an $2$-hop ego network which we use for our experiments. We use the sum of GloVe embeddings \citep{pennington2014glove} in the abstract as node features. For all datasets, we perform link prediction by training on a subset (i.e., a percentage) of the edges and then attempting to predict the unseen edges (with $20\%$ of the held-out edges used for validation).
}

\begin{table}[t]
\begin{center}
\begin{tabular}{llllllllll}
\toprule
           & \multicolumn{3}{c}{PPI}                          & \multicolumn{3}{c}{FirstMM DB}                   & \multicolumn{3}{c}{Ego-AMINER}                  \\ \midrule
Edges      & 10\%           & 20\%           & 30\%           & 10\%           & 20\%           & 30\%           & 10\%           & 20\%           & 30\%           \\ \bottomrule 
Meta-Graph & \textbf{0.795} & \textbf{0.833} & \textbf{0.845} & \textbf{0.782} & \textbf{0.786} & 0.783          & \textbf{0.626}          & \textbf{0.738} & \textbf{0.786} \\
MAML       & 0.770          & 0.815          & 0.828          & 0.776          & 0.782          & \textbf{0.793} & 0.561          & 0.662          & 0.667          \\
Random     & 0.578          & 0.651          & 0.697          & 0.742          & 0.732          & 0.720          & 0.500          & 0.500          & 0.500          \\
No Finetune & 0.738          & 0.786          & 0.801          & 0.740          & 0.710          & 0.734          & 0.548          & 0.621          & 0.673          \\
Finetune   & 0.752          & 0.801          & 0.821          & 0.752          & 0.735          & 0.723          & 0.623 & 0.691          & 0.723          \\
Adamic     & 0.540          & 0.623          & 0.697          & 0.504          & 0.519          & 0.544          & 0.515          & 0.549          & 0.597          \\
Deepwalk   & 0.664          & 0.673          & 0.694          & 0.487          & 0.473          & 0.510          & 0.602          & 0.638          & 0.672          \\ \bottomrule
\end{tabular}
\vspace{-5pt}
\caption{Convergence AUC results for different training edge splits.}
 \vspace{-10pt}
\label{tab:convergence}
\end{center}
\end{table}

\subsection{Results}

\xhdr{Q1: Overall Performance}
Table \ref{tab:convergence} shows the link prediction AUC for Meta-Graph and the baseline models when trained to convergence using 10\%, 20\% or 30\% of the graph edges. In this setting, we adapt the link prediction models on the test graphs until learning converges, as determined by performance on the validation set of edges, and we report the average link prediction AUC over the test edges of the test graphs. 
Overall, we find that Meta-Graph achieves the highest average AUC in all but one setting, with an average relative improvement of $4.8\%$ in AUC compared to the MAML approach and an improvement of $5.3\%$ compared to the Finetune baseline.
Notably, Meta-Graph is able to maintain especially strong performance when using only $10\%$ of the graph edges for training, highlighting how our framework can learn from very sparse samples of edges. Interestingly, in the Ego-AMINER dataset, unlike PPI and FIRSTMM DB, we observe the relative difference in performance between Meta-Graph and MAML to increase with density of the training set. We hypothesize that this is due to the fickle nature of optimization with higher order gradients in MAML \citep{antoniou2018train} which is somewhat alleviated in GS-gating due to the gating mechanism. With respect to computational complexity we observe only a slight overhead when comparing Meta-Graph to MAML, which can be reconciled by realizing that the graph signature function is not updated in the inner loop update but only in outer loop.
In the Appendix, we provide additional results when using larger sets of training edges, and, as expected, we find that the relative gains of Meta-Graph decrease as more and more training edges are available. 

\begin{table}[]
\begin{center}
\begin{tabular}{llllllllll}
\toprule
           & \multicolumn{3}{c}{PPI}                          & \multicolumn{3}{c}{FirstMM DB}                   & \multicolumn{3}{c}{Ego-AMINER}                   \\ \midrule
Edges      & 10\%           & 20\%           & 30\%           & 10\%           & 20\%           & 30\%           & 10\%           & 20\%           & 30\%           \\ \bottomrule
Meta-Graph & \textbf{0.795} & \textbf{0.824} & \textbf{0.847} & \textbf{0.773} & \textbf{0.767} & 0.737 & \textbf{0.620} & 0.585          & \textbf{0.732} \\
MAML       & 0.728          & 0.809          & 0.804          & 0.763          & 0.750          & \textbf{0.750}          & 0.500          & 0.504          & 0.500          \\
No Finetune & 0.600          & 0.697          & 0.717          & 0.708          & 0.680          & 0.709          & 0.500          & 0.500          & 0.500          \\
Finetune   & 0.582          & 0.727          & 0.774          & 0.705          & 0.695          & 0.704          & 0.608          & \textbf{0.675} & 0.713         \\ \bottomrule 
\end{tabular}
\vspace{-7pt}
\caption{5-gradient update AUC results with various fractions of training edges.}
\label{tab:fast}
\end{center}
\vspace{-5pt}
\end{table}

\begin{figure}[t!]
 \vspace{-15pt}
    \begin{center}
    \includegraphics[width=0.49\linewidth]{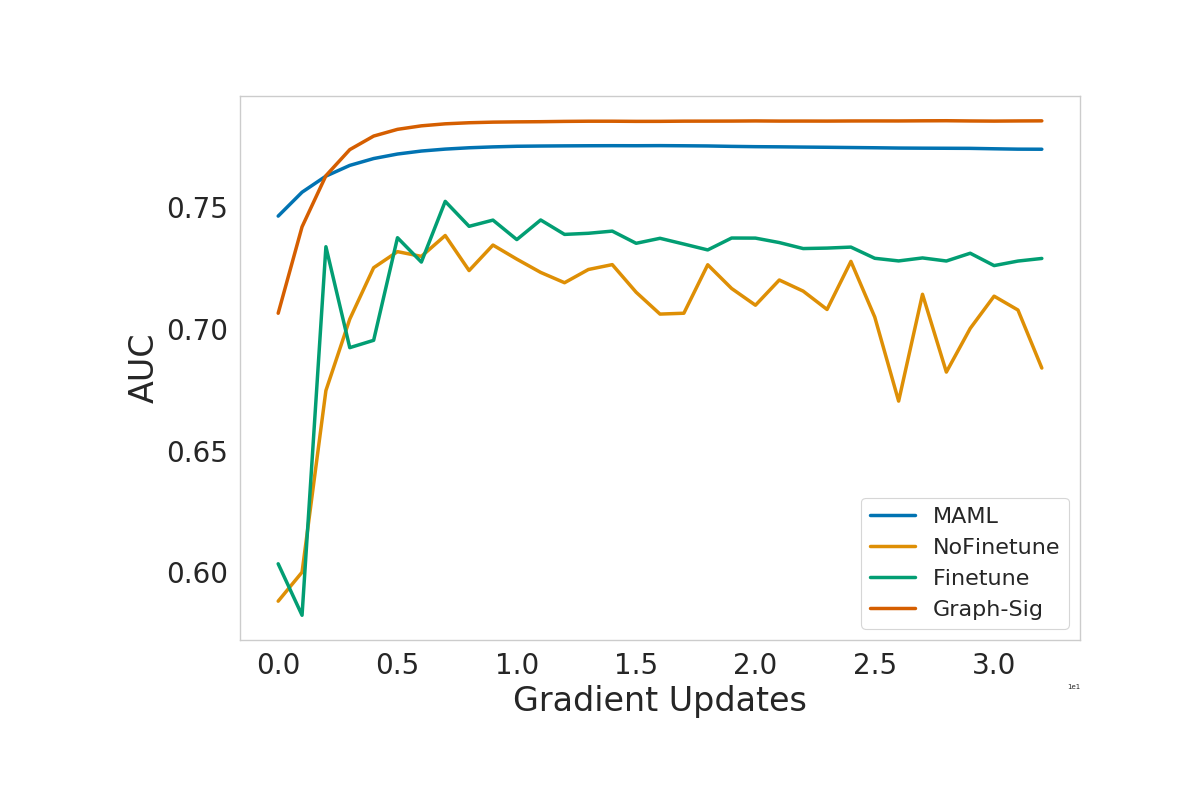}
    \includegraphics[width=0.49\linewidth]{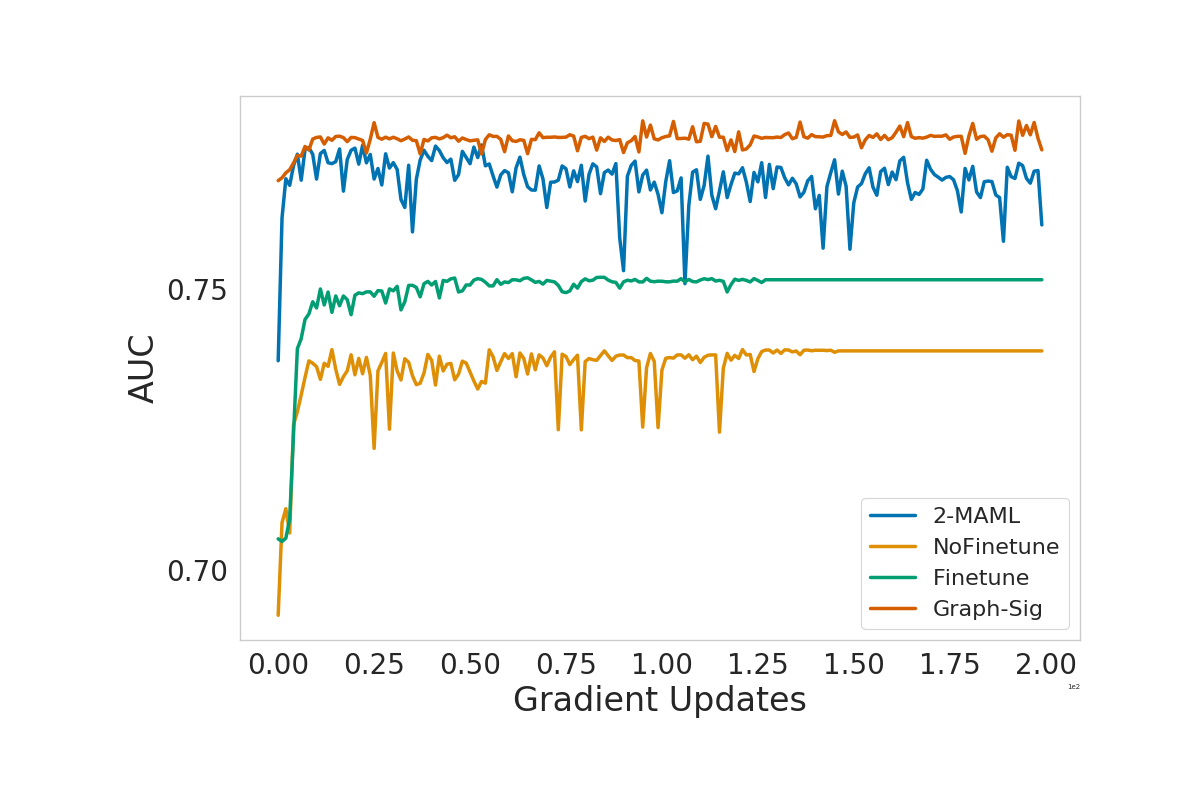}
    \vspace{-15pt}
    \caption{AUC scores on PPI \textbf{(Left)} and FirstMM DB \textbf{(Right)} graphs with $10\%$ of edges observed.}
    \label{fig:fast_adaptation}
    %\end{minipage}
    \end{center}
    \vspace{-10pt}
\end{figure}

\xhdr{Q2: Fast Adaptation}
Table \ref{tab:fast} highlights the average AUCs achieved by Meta-Graph and the baselines after performing only 5 gradient updates on the batch of training edges. 
Note that in this setting we only compare to the MAML, Finetune, and No Finetune baselines, as fast adaption in this setting is not well defined for the DeepWalk and Adamic-Adar baselines.
In terms of fast adaptation, we again find that Meta-Graph is able to outperform all the baselines in all but one setting, with an average relative improvement of $9.4\%$ compared to MAML and $8.0\%$ compared to the Finetune baseline---highlighting that Meta-Graph can  not only learn from sparse samples of edges but is also able to quickly learn on new data using only a small number of gradient steps. Also, we observe poor performance for MAML in the Ego-AMINER dataset dataset which we hypothesize is due to extremely low learning rates ---i.e. $1e-7$ needed for any learning, the addition of a graph signature alleviates this problem.
Figure \ref{fig:fast_adaptation} shows the learning curves for the various models on the PPI and FirstMM DB datasets, where we can see that Meta-Graph learns very quickly but can also begin to overfit after only a small number of gradient updates, making early stopping essential.

\begin{figure}[t!]
    \begin{center}
    %begin{minipage}{0.49\linewidth}
    \includegraphics[width=0.49\linewidth]{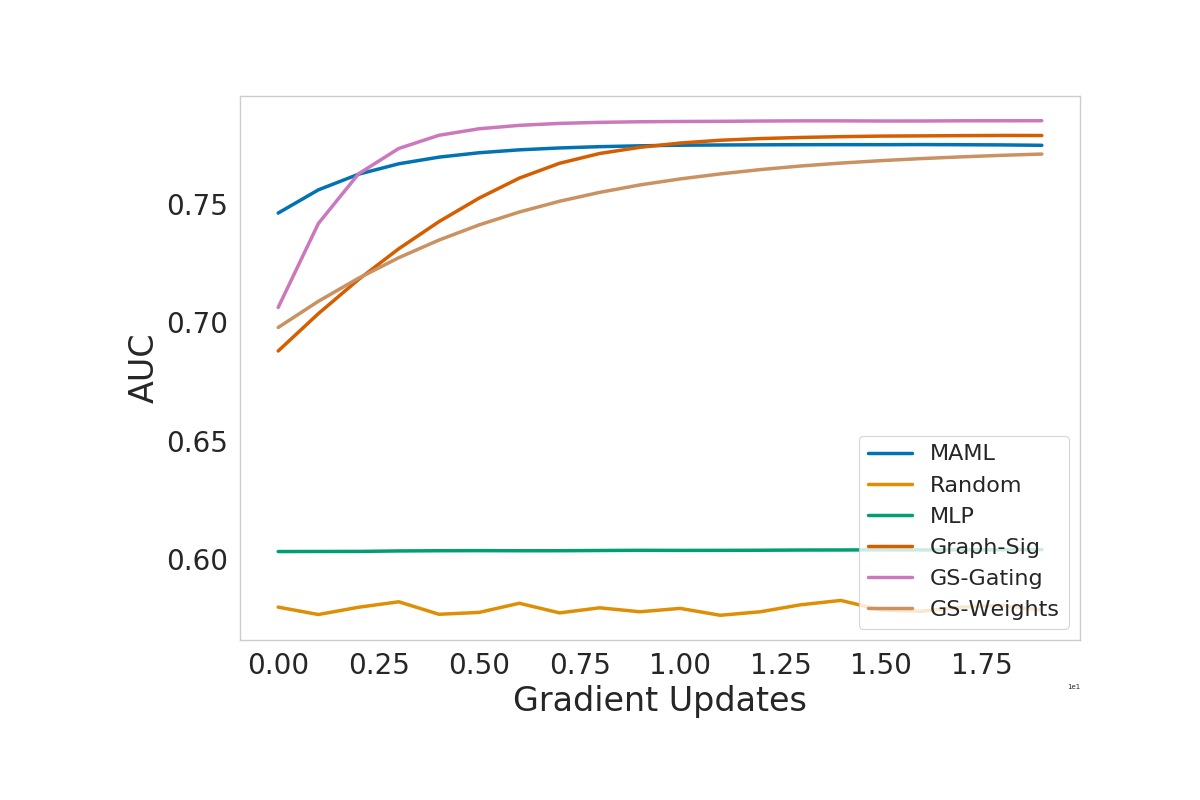}
    %\caption{Ablation of various Meta-Graph architecture on FirstMM DB graphs with $10\%$ of edges observed.}
    %\label{fig:ppi_ablation}
    %\end{minipage}
    %\begin{minipage}{0.49\linewidth}
     \includegraphics[width=0.49\linewidth]{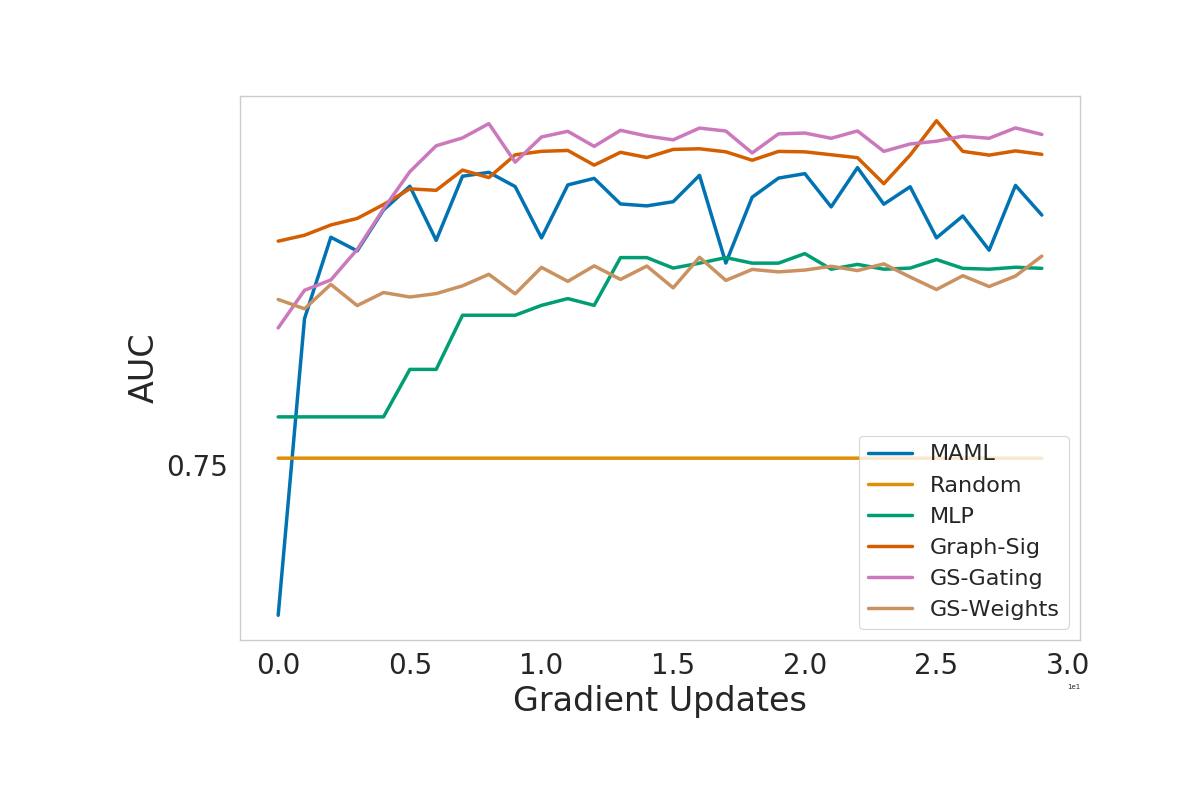}
     \vspace{-15pt}
     \caption{Ablation study on PPI \textbf{(Left)} and FirstMM DB \textbf{(Right)} graphs with $10\%$ of edges.}
      \label{fig:ablation}
    %\end{minipage}
    \end{center}
    \vspace{0pt}
\end{figure}

\xhdr{Q3: Choice of Meta-Graph Architecture}
We study the impact of the graph signature function and its variants GS-Gating and GS-Weights by performing an ablation study using the FirstMM DB dataset. Figure \ref{fig:ablation} shows the performance of the different model variants and baselines considered as the training progresses. In addition to models that utilize different signature functions we report a random baseline where parameters are initialized but never updated allowing us to assess the inherent power of the VGAE model for few-shot link prediction. To better understand the utility of using a GCN based inference network we also report a VGAE model that uses a simple MLP on the node features and is trained analogously to Meta-Graph as a baseline. As shown in Figure \ref{fig:ablation} many versions of the signature function start at a better initialization point or quickly achieve higher AUC scores in comparison to MAML and the other baselines, but simple modulation and GS-Gating are superior to GS-Weights after a few gradient steps.

\noindent\textbf{Q4: What is learned by the graph signature?}
To gain further insight into what knowledge is transferable among graphs we use the FirstMM DB and Ego-AMINER datasets to probe and compare the output of the signature function with various graph heuristics. In particular, we treat the output of $s_\G = \psi(\G)$ as a vector and compute the cosine similarity between all pairs of graph in the training set (i.e., we compute the pairwise cosine similarites between graph signatures, $s_\G$).
We similarly compute three pairwise graph statistics---namely, the cosine similarity between average node features in the graphs, the difference in number of nodes, and the difference in number of edges---and we compute the Pearson correlation between the pairwise graph signature similarities and these other pairwise statistics. 
As shown in Table \ref{Table:Analysis} we find strong positive correlation in terms of Pearson correlation coefficient between node features and the output of the signature function for both datasets, indicating that the graph signature function is highly sensitive to feature information. 
This observation is not entirely surprising given that we use such sparse samples of edges---meaning that many structural graph properties are likely lost and making the meta-learning heavily reliant on node feature information. We also observe moderate negative correlation with respect to the average difference in nodes and edges between pairs of graphs for FirstMM DB dataset. For Ego-AMINER we observe small positive correlation for difference in nodes and edges.

\begin{table}[]{\small
%\resizebox{\textwidth}{!} {
\begin{center}

\begin{tabular}{@{}lllllll@{}}

\toprule
& \multicolumn{3}{c}{FirstMM DB} & \multicolumn{3}{c}{Ego-AMINER} \\ \midrule
\% Edges  & 10\% & 20\% & 30\%  & 10\% & 20\%  & 30\% \\ \bottomrule
Node Feats & 0.928 & 0.950 & 0.761 & 0.473 & 0.385 & 0.448\\
Diff Num. Nodes &-0.093 & -0.196 & -0.286 & 0.095 & 0.086 & 0.085 \\
Diff Num. Edges &-0.093 &  -0.195 & -0.281 & 0.093 & 0.072 & 0.075\\
\bottomrule
\end{tabular}
\end{center}
%}
}
\vspace{-10pt}
\caption{Pearson scores between graph signature output and other graph statistics.}
\label{Table:Analysis}
\vspace{-10pt}
\end{table}

\cut{
%%%%%%%%% Full Results Below this Line %%%%%%%%%%%

\begin{table*}[ht]
\label{ml-table}
\begin{center}

\begin{tabular}{lccccccccr}
\toprule
PPI \\ Convergence  &  $10\%$ & $20\%$ &  $30\%$ & $40\%$  & $50\%$ & $60\%$ & $70\%$\\
\midrule
\textbf{Meta-Graph-Sig}    & 785/0.802 & \textbf{0.831/0.841} & \textbf{0.846/0.849} & 0.848/0.852 & 0.848/0.847& 0.853/0.855 & 0.855/0.855 \\
\textbf{Meta-Graph-NC}    & \textbf{0.795/0.813} & \textbf{0.833/0.839} & 0.845/0.846 & \textbf{0.853/0.854} & 0.845/0.850& 0.843/0.848 & 0.779/0.799 \\
MG-Random & 0.733/0.738 & 0.775/0.764 & 0.802/0.790 & 0.816/0.798& 0.828/0.812& 0.838/0.822 & 0.839/0.814 \\
MAML-Concat   & 0.745/0.733 & 0.820/0.832 & 0.840/0.846 & 0.852/0.853& \textbf{0.854/0.854} & 0.856/0.857 & 0.863/0.861 \\
\textbf{MAML}    & 0.770/0.785 &0.815/0.825 & 0.828/0.834 & 0.838/0.841& 0.843/0.845& 0.850/0.850 & \textbf{0.858/0.856} \\
Random    & 0.578/0.530 & 0.651/0.590 & 0.697/0.639 & 0.729/0.682& 0.756/0.714& 0.778/0.743 & 0.795/0.762  \\
\textbf{No Finetune}    &0.738/0.757 & 0.786/0.803 & 0.801/0.820 & 0.817/0.831 & 0.827/0.840 &0.837/0.851 & 0.836/0.852 \\
\textbf{Finetune}    & 0.752/0.759  &0.8010/0.817 &0.821/0.835 & 0.832/ 0.847& 0.818/0.837 &  \textbf{0.856/0.862} & 0.841/0.852 \\
Adamic    & 0.540/0.540 & 0.623/0.622 & 0.697/0.700 & 0.756/0.761& 0.796/0.802& 0.827/0.834 & 0.849/0.856 \\
MAML-MLP    & 0.603/0.604 & 0.606/0.607 & 0.606/0.603 & 0.606/0.603& 0.604/0.604& 0.604/0.603 & 0.605/0.607 \\
Deepwalk    & 0.664/0.641 & 0.673/0.669 & 0.694/0.701& 0.727/0.743& 0.731/0.747 & 0.747/0.767 & 0.761/0.783\\

\bottomrule
\end{tabular}
\end{center}
\vskip -0.1in
\end{table*}

\begin{table*}[ht]
\label{ml-table}
\begin{center}

\begin{tabular}{lccccccccr}
\toprule
PPI-5 updates  &  $10\%$ & $20\%$ &  $30\%$ & $40\%$  & $50\%$ & $60\%$ & $70\%$\\
\midrule
\textbf{Meta-Graph-Sig}& 0.742/0.718 & \textbf{0.829/0.838} & 0.846/0.847 & 0.847/0.849 &0.848/0.848 & \textbf{0.854/0.857} &0.847/0.846\\
\textbf{Meta-Graph-NC}& \textbf{0.795/0.812} & 0.824/0.832 & \textbf{0.847/0.849} & \textbf{0.853/0.853} & 0.841/0.847& 0.840/0.845 & 0.847/0.840 \\
MAML-Concat   & 0.756/0.756 & 0.837/0.841 & 0.840/0.846 & 0.852/0.855& \textbf{0.855/0.855} & \textbf{0.855/0.85}5 & \textbf{0.856/0.855} \\
\textbf{MAML}    & 0.728/0.710 &0.809/0.816 & 0.804/0.806 & 0.825/0.829& 0.834/0.841& 0.846/0.847 & 0.851/0.852 \\
\textbf{No Finetune} & 0.600/0.547 & 0.697/0.668 &0.717/0.671 & 0.784/0.781 & 0.814/0.811 & 0.779/0.746& 0.822/0.821 \\
\textbf{Finetune} & 0.582/0.546 & 0.727/0.733 &0.774/0.788 & 0.702/0.722 & 0.804/0.826 & 0.718/0.716 & 0.766/0.777 \\
MAML-MLP    & 0.603/0.604 & 0.606/0.607 & 0.603/0.602 & 0.604/0.603& 0.603/0.602& 0.606/0.605 & 0.605/0.607 \\

\bottomrule
\end{tabular}
\end{center}
\vskip -0.1in
\end{table*}

\begin{table*}[ht]
\label{ml-table}
\begin{center}
\begin{tabular}{lccccccccr}
\toprule
FirstMM DB\\
Convergence  &  $10\%$ & $20\%$ &  $30\%$ & $40\%$  & $50\%$ & $60\%$ & $70\%$\\
\midrule
\textbf{Meta-Graph-Sig} & 0.779/0.713 & 0.781
0.718 & 0.771/0.695 &0.764/0.693 & 0.755/0.674 & 0.745/0.663 & 0.726/0.637\\
MG-Random & 0.557/0.569 & 0.507/0.506 & 0.477/0.477 & 0.444/0.454& 0.412/0.438& 0.383/0.423 & 0.358/0.409 \\
\textbf{Meta-Graph-NC} & \textbf{0.782/0.715} & \textbf{0.786/0.718} &0.783/0.712  & \textbf{0.781/0.707} & 0.760/0.670 &0.746/0.667 & 0.739/0.660 \\
MG-Random & 0.557/0.569 & 0.507/0.506 & 0.477/0.477 & 0.444/0.454& 0.412/0.438& 0.383/0.423 & 0.358/0.409 \\
\textbf{MAML}    & 0.776/0.712 &0.782/0.713 & \textbf{0.793/0.733} & \textbf{0.785/0.703} & \textbf{0.791/0.722}& 0.663/0.590 & 0.788/0.720 \\
Random    & 0.742/0.677 & 0.732/0.665 & 0.720/0.649& 0.714/0.638 & 0.705/0.629& 0.698/0.622 & 0.695/0.672  \\
\textbf{No Finetune} &0.740/0.692 & 0.710/0.646 & 0.734/0.687& 0.722/0.682 & 0.712/0.671  & 0.710/0.664 & 0.698/0.657\\
\textbf{Finetune}    & 0.752/0.701  & 0.735/0.690  & 0.723/0.672 &0.734/0.688 & 0.749/0.719  & 0.700/0.662 & 0.695/0.657 \\
Adamic    & 0.504/0.504 &0.519/0.519 & 0.544/0.543 & 0.573/0.569 & 0.604/0.602 & 0.643/0.636 &0.678/0.676\\
MAML-MLP    &0.774/0.697 & 0.759/0.684 &0.766/0.691 &0.765/0.689 & 0.789/0.714 & 0.779/0.703 & 0.801/0.726 \\
Deepwalk    & 0.487/0.492 & 0.473/0.525 & 0.510/0.604& 0.608/0.719& 0.722/0.798 & \textbf{0.832/0.882} & \textbf{0.911/0.923} \\

\bottomrule
\end{tabular}
\end{center}
\vskip -0.1in
\end{table*}

\begin{table*}[ht]
\label{ml-table}
\begin{center}
\begin{tabular}{lccccccccr}
\toprule
FirstMM DB \\
5 updates  &  $10\%$ & $20\%$ &  $30\%$ & $40\%$  & $50\%$ & $60\%$ & $70\%$\\
\midrule
\textbf{Meta-Graph-Sig} & 0.770/0.712  & 0.764/0.714 & 0.743/0.677 & 0.706/0.643 & 0.742/0.676&0.732/0.663 &0.688/0.635 \\
\textbf{Meta-Graph-NC} & \textbf{0.773/0.713}  & \textbf{0.767/0.715} & 0.737/0.667 & \textbf{0.759/0.697} & 0.740/0.653 & 0.737/0.667 & 0.679/0.632\\
\textbf{MAML} & 0.763/0.702 & 0.750/0.672 & 0.624/ 0.590 & \textbf{0.776/0.689} & 0.759/0.706  & 0.663/0.590 & 0.738/0.641\\
\textbf{No Finetune} & 0.708/0.644 & 0.680/0.603 & 0.709/0.643 & 0.701/0.646 &0.685/0.633 &0.683/0.628 & 0.653/0.598\\
\textbf{Finetune} & 0.705/0.655 & 0.695/0.646 & 0.704/0.633 & 0.704/0.653 &0.696/0.646 &0.658/0.624 & 0.670/0.637\\
MAML-MLP & 0.754 / 0.680 & 0.754/0.680 & \textbf{0.758/0.688} & 0.755/0.681 & \textbf{0.764/0.694}& \textbf{0.762/0.691} & \textbf{0.769 / 0.699} \\

\bottomrule
\end{tabular}
\end{center}
\vskip -0.1in
\end{table*}

\begin{table*}[ht]
\label{ml-table}
\begin{center}
\begin{tabular}{lccccccccr}
\toprule
Aminer-20k\\
Convergence  &  $10\%$ & $20\%$ &  $30\%$ & $40\%$  & $50\%$ & $60\%$ & $70\%$\\
\midrule
\textbf{Meta-Graph} & 0.681/0.656 & 0.738/0.719 & 0.790/0.775 & 0.825/0.813 & 0.847/0.834 & 0.863/0.855& 0.856/0.843\\
MG-Random &  0.500/0.500 & 0.500/0.500 & 0.500/0.500 & 0.500/0.500 & 0.500/0.500 & 0.500/0.500 & 0.500/0.500\\
\textbf{MAML}    & 0.690/0.660 & 0.751/0.729 & 0.793/0.757 & 0.831/0.805 & 0.828/0.804& 0.861/0.841 &0.857/0.840\\
Random    &  0.500/0.500 & 0.500/0.500 & 0.500/0.500 & 0.500/0.500 & 0.500/0.500 & 0.500/0.500 & 0.500/0.500\\
\textbf{No Finetune} & 0.675/0.657& 0.788/0.776&0.834/0.831& 0.859/0.853 &0.890/0.891 &0.877/0.867 & 0.835/0.833\\
\textbf{Finetune} & 0.812/0.800 & 0.879/0.873 & 0.890/0.881 & 0.905/0.902 & 0.912/0.910 & 0.931/0.929 & 0.942/0.942\\
Adamic    & 0.514/0.514 & 0.553/0.553 & 0.606/0.607 & 0.664/0.664 & 0.720/0.721 &0.772/0.773 &0.814/0.815\\
MAML-MLP    &0.701/0.696 &0.711/0.706 &0.703/0.6983 &0.714/0.707 &0.710/0.704 &0.712/0.703 & 0.711/0.705 \\
Deepwalk    & 0.588/0.632 & 0.714/0.781 & 0.829/0.873& 0.898/0.921& 0.931/0.944 & 0.949/0.956 & 0.960/0.964 \\

\bottomrule
\end{tabular}
\end{center}
\vskip -0.1in
\end{table*}

\begin{table*}[ht]
\label{ml-table}
\begin{center}
\begin{tabular}{lccccccccr}
\toprule
Aminer-20k \\
5 updates  &  $10\%$ & $20\%$ &  $30\%$ & $40\%$  & $50\%$ & $60\%$ & $70\%$\\
\midrule
\textbf{Meta-Graph} & \\
\textbf{MAML} & \\
\textbf{No Finetune} & 0.500/0.500 & 0.500/0.500 & 0.500/0.500 & 0.500/0.500 & 0.500/0.500 &  0.500/0.500 & 0.500/0.500\\
\textbf{Finetune} & 0.702/0.684 & 0.821/0.811 & 0.847/0.838& 0.853/0.850 & 0.884 & 0.858/0.854 & 0.886/0.884\\
MAML-MLP &  0.601/0.595 & 0.597/0.59 & 0.601/0.592& 0.613/0.612 & 0.878 & 0.614/0.608 & 0.630/0.623\\

\bottomrule
\end{tabular}
\end{center}
\vskip -0.1in
\end{table*}

\cut{
\begin{table*}[ht]
\label{ml-table}
\begin{center}
\begin{tabular}{lccccccccr}
\toprule
Ego-AMINER\\
Convergence  &  $10\%$ & $20\%$ &  $30\%$ & $40\%$  & $50\%$ & $60\%$ & $70\%$\\
\midrule
\textbf{Meta-Graph-Sig} & 0.616/0.582 & 0.712/0.694& 0.739/0.754 & 0.791/0.779 & 0.792/0.776 & 0.817/0.806 & 0.786/0.780\\
\textbf{Meta-Graph} & 0.607/0.574 & 0.606/0.656& 0.781/0.780& 0.685/0.659 & 0.771/0.788 & 0.794/0.793 & 0.778/0.775\\
MG-Random & 0.500/0.500 & 0.500/0.500 & 0.500/0.500 & 0.500/0.500 & 0.500/0.500 & 0.500/0.500 & 0.500/0.500\\
\textbf{MAML}    & 0.561/0.540 & 0.662/0.649 & 0.667/0.669 & 0.682/0.676 & 0.720/0.708 & 0.741/0.759  & 0.768/0.774\\
Random    & 0.500/0.500 & 0.500/0.500 & 0.500/0.500 & 0.500/0.500 & 0.500/0.500 & 0.500/0.500 & 0.500/0.500 \\
\textbf{No Finetune} & 0.548/0.532 & 0.621/0.653 & 0.673/0.708 & 0.702/0.724 & 0.652/0.636 & 0.745/0.778 & 0.769/0.792\\
\textbf{Finetune} & 0.623/0.637& 0.691/0.705 & 0.723/0.745 & 0.764/0.782 & 0.767/0.793 & 0.792/0.814 & 0.781/0.805\\
Adamic    &0.515/0.516 & 0.549/0.553& 0.597/0.605& 0.655/0.669& 0.693/0.703& 0.744/0.758&0.772/0.793\\
MAML-MLP    & \\
Deepwalk    &0.602/0.620& 0.638/0.642 &0.672/0.669& 0.686/0.685 &0.689/0.679 & 0.711/0.706&0.731/0.714\\

\bottomrule
\end{tabular}
\end{center}
\vskip -0.1in
\end{table*}
}

\begin{table*}[ht]
\label{ml-table}
\begin{center}
\begin{tabular}{lccccccccr}
\toprule
Ego-AMINER \\
5 updates  &  $10\%$ & $20\%$ &  $30\%$ & $40\%$  & $50\%$ & $60\%$ & $70\%$\\
\midrule
\textbf{Meta-Graph-Sig}&0.606/0.572 & 0.500/0.500 & 0.737/0.741 & 0.500/0.500 & 0.500/0.500 &0.500/0.500 & 0.500/0.500\\
\textbf{Meta-Graph} &0.501/0.501& 0.655/0.685& 0.755/0.732& 0.500/0.500 &0.790/0.787 &0.733/0.710&0.500/0.500\\
\textbf{MAML} & 0.500/0.500 & 0.504/0.502 & 0.500/0.500& 0.500/0.500 & 0.519/0.510 & 0.500/0.500 & 0.500/0.500\\
\textbf{No Finetune} & 0.500/0.500 & 0.500/0.500 & 0.500/0.500 & 0.500/0.500 & 0.500/0.500 & 0.500/0.500 & 0.500/0.500\\
\textbf{Finetune} & 0.608/0.589 & 0.675/0.668 & 0.713/0.720 & 0.755/0.767 & 0.744/0.758 & 0.706/0.734 & 0.671/0.720\\
MAML-MLP &  \\

\bottomrule
\end{tabular}
\end{center}
\vskip -0.1in
\end{table*}

\begin{table*}[ht]
\label{ml-table}
\begin{center}
\begin{tabular}{lccccccccr}
\toprule
PPI 
  &  $10\%$ & $20\%$ &  $30\%$ & \textbf{$40\%$-high p-val}& $50\%$ & $60\%$ & $70\%$\\
\midrule
Log Degree \\ Difference & NaN& 0.586/0.807& 0.493/0.699 & -0.063/-0.045 & -0.364/-0.368 &-0.638/-0.688 & -0.796/-0.851\\
Avg-Emb &NaN & -0.508/-0.891& -0.469/-0.847 & 0.008/-0.018 & 0.298/0.186 & 0.555/0.522& 0.860/0.660 \\
WL-kernel &NaN & -0.298/-0.284& -0.299/-0.276& -0.155/-0.132 & \textbf{high p-val}& \textbf{high p-val} & 0.342/0.245\\

\bottomrule
\end{tabular}
\end{center}
\vskip -0.1in
\end{table*}
}

\cut{
We evaluate Meta-Graph against multiple baselines for both final convergence as well as fast adaptation using the AUC classification accuracy and Average Precision for predicting real vs.\@ randomly sampled negative edges. We report the best variant of Meta-Graph using as determined by the validation set and in section \ref{ablation_section} we perform an ablation among the different architecture choices. Table \ref{Table:PPI_FIRST_convergence}. shows convergence results for both PPI and FirstMM DB datasets and Table \ref{Table:Ego_Convergence} for Ego-AMINER citation networks under different training edge splits used during meta-training. When testing for model convergence we adapt to new test graphs until learning converges as determined by performance on the validation set of edges. In addition, we also report in Table \ref{Table:PPI_FIRST_fast_adaptation} and Table. \ref{Table:Ego_fast_adaptation} results in the fast adaptation setting where each approach has $5$-gradient steps to quickly adapt to each new graph in the test set. 

We compare Meta-Graph against a number of classical link prediction baselines: Adamic-Adar \citep{adamic2003friends}, DeepWalk \citep{perozzi2014deepwalk} and a GCN model with random weights to understand the natural expressive power of the base VGAE model. We also report a NoFinetuning and Finetuning baselines. The former trains a single set of VGAE parameters for each graph, as a result, each model is independent of the other graphs in the dataset. For Finetuning the graphs are observed in a sequential order and the weights are finetuned starting from the previous graph in the sequence. Finally, we also compare with a meta-learning baseline which does not include GS, which we call MAML as there are both global and local parameters. We implement all of our models using the Pytorch-Geometric libarary \citep{Fey/Lenssen/2019} and for fair comparison we tune all learning rates and meta-learning specific hyperparameters such as the number of local updates using Bayesian Optimization with Thompson sampling \citep{kandasamy2018parallelised} on the validation set. 

As shown in Table \ref{Table:PPI_FIRST_convergence} we find that Meta-Graph outperforms all baselines for PPI in terms of final convergence by a significant margin for all training edge splits. A similar result is observed for FirstMM DB for $10\%$ and $20\%$ of edges, while for $30\%$ MAML (which itself is a meta-learning algorithm) marginally outperforms Meta-Graph. While, for convergence results on the Ego-AMINER dataset we observe that simply applying MAML is not sufficient to beat a strong Finetuning baseline, but instead Meta-Graph is able to outperform all baselines suggesting that the signature function plays a crucial role. As meta-learning approaches are purpose built for fast adaptation we find that both Meta-Graph and MAML achieve large performance gains from just $5$ gradient updates compared to all other baseline on both PPI and FirstMM DB as shown in Fig. \ref{fig:ppi_fast} and Fig. \ref{fig:FirstMM_fast} in the extreme sparse setting with only $10\%$ of training edges. In Table \ref{Table:PPI_FIRST_fast_adaptation} and \ref{Table:Ego_fast_adaptation} we report fast adaptation results for each dataset under the same settings as the convergence results. We find that the addition of the graph signature function in Meta-Graph further boosts performance by learning a better initialization point relative to MAML across all datasets and is superior to all other baselines aside from a strong Finetuning baseline om the Ego-AMINER dataset with $20\%$ of training edges.}

\section{Related Work}
We now briefly highlight related work on link prediction, meta-learning, few-shot classification, and few-shot learning in knowledge graphs. Link prediction considers the problem of predicting missing edges between two nodes in a graph that are likely to have an edge. \citep{Liben-Nowell:2003:LPP:956863.956972}. Common successful applications of link prediction include friend and content recommendations \citep{Aiello:2012:FPH:2180861.2180866}, shopping and movie recommendation \citep{Huang:2005:LPA:1065385.1065415}, knowledge graph completion \citep{nickel2015review} and even important social causes such as identifying criminals based on past activities \citep{Hasan06linkprediction}. Historically, link prediction methods have utilized topological graph features (e.g., neighborhood overlap), yielding strong baselines such as the Adamic/Adar measure \citep{adamic2003friends}. Other approaches include matrix factorization methods \citep{Menon:2011:LPV:2034117.2034146} and more recently deep learning and graph neural network based approaches \citep{grover2016node2vec,wang2015link,zhang2018link}. A commonality among all the above approaches is that the link prediction problem is defined over a single dense graph where the objective is to predict unknown/future links within the same graph. Unlike these previous approaches, our approach considers link prediction tasks over multiple sparse graphs that are drawn from distribution over graphs akin to real world scenario such as protein-protein interaction graphs, 3D point cloud data and citation graphs in different communities. 

In meta-learning or learning to learn \citep{bengio1990learning,bengio1992optimization,thrun2012learning,schmidhuber1987evolutionary}, the objective is to learn from prior experiences to form inductive biases for fast adaptation to unseen tasks. Meta-learning has been particularly effective in few-shot learning tasks with a few notable approaches broadly classified into metric based approaches \citep{vinyals2016matching,snell2017prototypical,koch2015siamese}, augmented memory \citep{santoro2016meta,kaiser2017learning,mishra2017simple} and optimization based approaches \citep{finn2017model,lee2018gradient}. Recently, there are several works that lie at the intersection of meta-learning for few-shot classification and graph based learning. In Latent Embedding Optimization, \citet{rusu2018meta} learn a graph between tasks in embedding space while \cite{liu2019propagate} introduce a message propagation rule between prototypes of classes. However, both these methods are restricted to the image domain and do not consider meta-learning over a distribution of graphs as done here.

Another related line of work considers the task of few-shot relation prediction in knowledge graphs. \citet{xiong2018one} developed the first method for this task, which leverages a learned matching metric using
both a learned embedding and one-hop graph structures. More recently \citet{chen2019meta} introduce Meta Relational Learning framework (MetaR) that seeks to transfer relation-specific meta information to new relation types in the knowledge graph. A key distinction between few-shot relation setting and the one which we consider in this work is that we assume a distribution over graphs, while in the knowledge graph setting there is only a single graph and the challenge is generalizing to new types of relations within this graph.

\cut{
In meta-learning or learning to learn \citep{bengio1990learning,bengio1992optimization,thrun2012learning,schmidhuber1987evolutionary}, the objective is to learn from prior experiences to form inductive biases for fast adaptation to unseen tasks. While there are many approaches to meta-learning, in this work we focus on a class of approaches known as gradient-based meta-learning, where stochastic gradient descent is used to backpropagate through the learning process itself. Let $\mathcal{D}$ be a dataset which defines a distribution over tasks $\mathcal{T}$ with some shared structure and, each task $\mathcal{T}_j$ defines a distribution over datapoints, $x_j$, which is to be optimized by the meta-learner to produce task specific parameters. In the few-shot learning setting, the meta-learner observes up to $N$ samples from each task ---i.e. $x_{j_1}, ... , x_{j_N} \sim p(\mathcal{T}_{j})$ with the goal of finding a set of shared parameters, $\theta$, over tasks. These global parameters are optimized such that when a few gradient descent steps are taken from the initialization, $\theta$, given a small sample of points from $\mathcal{T}_{j}$ there is good generalization performance on held out samples also from the same task---i.e., $x_{j_{N+1}}, ... , x_{j_{N+m}} \sim p(\mathcal{T}_{j_i})$. That is, starting from the global parameters $\theta$ the meta-learning algorithm produces a set of local parameters, $\phi_j$ tailored to $\mathcal{T}_j$, through fast adaptation. This approach to gradient based meta-learning is known as Model Agonistic Meta-Learning (MAML) \citep{finn2017model}. 
}
\cut{
\xhdr{Link Prediction}
Link prediction (LP) is a problem of predicting whether two nodes are likely to have an edge in the graph \citep{Liben-Nowell:2003:LPP:956863.956972}. LP has been used in many applications like friend and content recommendations \citep{Aiello:2012:FPH:2180861.2180866} , shopping and movie recommendation \citep{Huang:2005:LPA:1065385.1065415} , knowledge graph completion \citep{nickel2015review} and even to identify criminals based on their activities \citep{Hasan06linkprediction}. Historically, link prediction has been mainly performed using topological features of the graph like common neighbors. This has yielded very strong baselines like Adamic/Adar measure \citep{adamic2003friends}, Jaccard Index among others. Other approaches include Matrix Factorization \citep{Menon:2011:LPV:2034117.2034146} and more recently graph neural networks \citep{zhang2018link} have been also been explored for this application. However, all of the above mentioned approaches dealt with a single dense graph where you try to predict unknown/future links. Unlike these previous approaches, our approach considers link prediction tasks on multiple sparse graphs which are drawn from similar real world scenario. 
}
\section{Discussion and Conclusion}
We introduce the problem of few-shot link prediction---where the goal is to learn from multiple graph datasets to perform link prediction using small samples of graph data---and we develop the Meta-Graph framework to address this task. 
Our framework adapts gradient-based meta learning to optimize a shared parameter initialization for local link prediction models, while also learning a parametric encoding, or signature, of each graph, which can be used to modulate this parameter initialization in a graph-specific way. Empirically, we observed substantial gains using Meta-Graph compared to strong baselines on three distinct few-shot link prediction benchmarks. 
In terms of limitations and directions for future work, one key limitation is that our graph signature function is limited to modulating the local link prediction model through an encoding of the current graph, which does not explicitly capture the pairwise similarity between graphs in the dataset. 
Extending Meta-Graph by learning a similarity metric or kernel between graphs---which could then be used to condition meta-learning---is a natural direction for future work. Another interesting direction for future work is extending the Meta-Graph approach to multi-relational data, and exploiting similarities between relation types through a suitable graph signature function.

%A similar result is also observed on one popular approach to gradient based meta-learning in MAML for both final convergence but critically fast adaptation in few gradient steps. %While, the GS uses an element-wise modulation of the GCN weights it is not the only possible choice. Extending Meta-Graph with different mechanisms for parameter modulation and gaining deeper insight over the learned parameters in GS is a fruitful direction for future work.

\subsection*{Acknowledgements}
The authors would like to thank Thang Bui, Maxime Wabartha, Nadeem Ward, Sebastien Lachapelle, and Zhaocheng Zhu for helpful feedback on earlier drafts of this work. In addition, the authors would like to thank the Uber AI team including other interns that helped shape earlier versions of this idea. Joey Bose is supported by the IVADO PhD fellowship and this work was done as part of his internship at Uber AI.
\clearpage

\bibliography{bibliography.bib}
\bibliographystyle{iclr2020_conference}
\clearpage
\section{Appendix}
\subsection{A: Ego-Aminer Dataset Construction}
To construct the Ego-Aminer dataset we first create citation graphs from different fields of study. We then select the top $100$ graphs in terms number of nodes for further pre-processing. Specifically, we take the $5$-core of each graph ensuring that each node has a minimum of $5$-edges. We then construct ego networks by randomly sampling a node from the $5$-core graph and taking its two hop neighborhood. Finally, we remove graphs with fewer than $100$ nodes and greater than $20000$ nodes which leads to a total of $72$ graphs as reported in Table \ref{datasetstats-table}.

\subsection{B: Additional Results}
We list out complete results when using larger sets of training edges for PPI, FIRSTMM DB and Ego-Aminer datasets. We show the results for two metrics i.e. Average AUC across all test graphs. As expected, we find that the relative gains of Meta-Graph decrease as more and more training edges are available.

\begin{table*}[ht]
\small
%\label{ml-table}
\begin{center}

\begin{tabular}{lccccccccr}
\toprule
PPI \\ Convergence  &  $10\%$ & $20\%$ &  $30\%$ & $40\%$  & $50\%$ & $60\%$ & $70\%$\\
\midrule
Meta-Graph    & \textbf{0.795} & \textbf{0.831} & \textbf{0.846} & \textbf{0.853} & 0.848& 0.853 & 0.855 \\
MAML   & 0.745 & 0.820 & 0.840 & 0.852& \textbf{0.854} & 0.856 & \textbf{0.863} \\
Random    & 0.578 & 0.651 & 0.697 & 0.729& 0.756& 0.778 & 0.795  \\
No Finetune    &0.738 & 0.786 & 0.801 & 0.817 & 0.827 &0.837 & 0.836 \\
Finetune    & 0.752  &0.8010 &0.821 & 0.832& 0.818 &  \textbf{0.856} & 0.841 \\
Adamic    & 0.540 & 0.623 & 0.697 & 0.756& 0.796& 0.827 & 0.849 \\
MAML-MLP    & 0.603 & 0.606 & 0.606 & 0.606& 0.604& 0.604 & 0.605 \\
Deepwalk    & 0.664 & 0.673 & 0.694& 0.727& 0.731 & 0.747 & 0.761\\

\bottomrule
\end{tabular}
\caption{AUC Convergence results for PPI dataset for training edge splits}
\end{center}
\vskip -0.1in

\end{table*}

\begin{table*}[ht]
\small
%\label{ml-table}
\begin{center}

\begin{tabular}{lccccccccr}
\toprule
PPI-5 updates  &  $10\%$ & $20\%$ &  $30\%$ & $40\%$  & $50\%$ & $60\%$ & $70\%$\\
\midrule
Meta-Graph & \textbf{0.795} & \textbf{0.829} & \textbf{0.847} & \textbf{0.853} &0.848 & \textbf{0.854} & \textbf{0.856}\\
MAML   & 0.756 & 0.837 & 0.840 & 0.852& \textbf{0.855} & \textbf{0.855} & \textbf{0.856} \\
%MAML    & 0.728/0.710 &0.809/0.816 & 0.804/0.806 & 0.825/0.829& 0.834/0.841& 0.846/0.847 & 0.851/0.852 \\
No Finetune & 0.600 & 0.697 &0.717 & 0.784 & 0.814 & 0.779& 0.822 \\
Finetune & 0.582 & 0.727 &0.774 & 0.702 & 0.804 & 0.718 & 0.766 \\
MAML-MLP    & 0.603 & 0.606 & 0.603 & 0.604& 0.603& 0.606 & 0.605 \\

\bottomrule
\end{tabular}
\caption{5-gradient update AUC results for PPI for training edge splits}
\end{center}
\vskip -0.1in
\end{table*}

\begin{table*}[ht]
\small
%\label{ml-table}
\begin{center}
\begin{tabular}{lccccccccr}
\toprule
FirstMM DB\\
Convergence  &  $10\%$ & $20\%$ &  $30\%$ & $40\%$  & $50\%$ & $60\%$ & $70\%$\\
\midrule
Meta-Graph & \textbf{0.782} & \textbf{0.786} &0.783  & \textbf{0.781} & 0.760 &0.746 & 0.739 \\
MAML    & 0.776 &0.782 & \textbf{0.793} & \textbf{0.785} & \textbf{0.791}& 0.663 & 0.788 \\
Random    & 0.742 & 0.732 & 0.720& 0.714 & 0.705& 0.698 & 0.695  \\
No Finetune &0.740 & 0.710 & 0.734& 0.722 & 0.712  & 0.710 & 0.698\\
Finetune    & 0.752  & 0.735  & 0.723 &0.734 & 0.749  & 0.700 & 0.695 \\
Adamic    & 0.504 &0.519 & 0.544 & 0.573 & 0.604 & 0.643 &0.678\\
%MAML-MLP    &0.774/0.697 & 0.759/0.684 &0.766/0.691 &0.765/0.689 & 0.789/0.714 & 0.779/0.703 & 0.801/0.726 \\
Deepwalk    & 0.487 & 0.473 & 0.510& 0.608& 0.722 & \textbf{0.832} & \textbf{0.911} \\
\bottomrule
\end{tabular}
\caption{AUC Convergence results for FIRSTMM DB dataset for training edge splits}
\end{center}
\vskip -0.1in
\end{table*}
\begin{table*}[ht]
\small
%\label{ml-table}
\begin{center}
\begin{tabular}{lccccccccr}
\toprule
FirstMM DB \\
5 updates  &  $10\%$ & $20\%$ &  $30\%$ & $40\%$  & $50\%$ & $60\%$ & $70\%$\\
\midrule
Meta-Graph & \textbf{0.773}  & \textbf{0.767} & \textbf{0.743} & \textbf{0.759} & 0.742& \textbf{0.732} &0.688 \\
MAML & 0.763 & 0.750 & 0.624 & \textbf{0.776} & \textbf{0.759}  & 0.663 & \textbf{0.738} \\
No Finetune & 0.708 & 0.680 & 0.709 & 0.701 &0.685 &0.683 & 0.653\\
Finetune & 0.705 & 0.695 & 0.704 & 0.704 &0.696 &0.658 & 0.670\\
%MAML-MLP & 0.754 / 0.680 & 0.754/0.680 & \textbf{0.758/0.688} & 0.755/0.681 & \textbf{0.764/0.694}& \textbf{0.762/0.691} & \textbf{0.769 / 0.699} \\
\bottomrule
\end{tabular}
\caption{5-gradient update AUC results for FIRSTMM DB for training edge splits}
\end{center}
\vskip -0.1in
\end{table*}
\begin{table*}[ht]
\small
%\label{ml-table}
\begin{center}
\begin{tabular}{lccccccccr}
\toprule
Ego-Aminer\\
Convergence  &  $10\%$ & $20\%$ &  $30\%$ & $40\%$  & $50\%$ & $60\%$ & $70\%$\\
\midrule
Meta-Graph & \textbf{0.626} & \textbf{0.738} & \textbf{0.786} & \textbf{0.791} & \textbf{0.792} & \textbf{0.817} & \textbf{0.786}\\
MAML    & 0.561 & 0.662 & 0.667 & 0.682 & 0.720 & 0.741 & 0.768\\
Random    & 0.500 & 0.500 & 0.500 & 0.500 & 0.500 & 0.500 & 0.500 \\
No Finetune & 0.548 & 0.621 & 0.673 & 0.702 & 0.652 & 0.7458 & 0.769\\
Finetune & 0.623& 0.691 & 0.723 & 0.764 & 0.767 & 0.792 & 0.781\\
Adamic    &0.515 & 0.549& 0.597& 0.655& 0.693& 0.744&0.772\\
Deepwalk    &0.602& 0.638 &0.672& 0.686 &0.689 & 0.711&0.731\\
\bottomrule
\end{tabular}
\caption{AUC Convergence results for Ego-Aminer dataset for training edge splits}
\end{center}
\vskip -0.1in
\end{table*}
\begin{table*}[ht]
\small
%\label{ml-table}
\begin{center}
\begin{tabular}{lccccccccr}
\toprule
Ego-Aminer \\
5 updates  &  $10\%$ & $20\%$ &  $30\%$ & $40\%$  & $50\%$ & $60\%$ & $70\%$\\
\midrule
Meta-Graph& \textbf{0.620} & 0.5850 & \textbf{0.732} & 0.500 & \textbf{0.790} & \textbf{0.733} & 0.500\\
MAML & 0.500 & 0.504 & 0.500& 0.500 & 0.519 & 0.500 & 0.500\\
No Finetune & 0.500 & 0.500 & 0.500 & 0.500 & 0.500 & 0.500 & 0.500\\
Finetune & 0.608 & \textbf{0.675} & 0.713 & \textbf{0.755} & 0.744 & 0.706 & \textbf{0.671} \\
\bottomrule
\end{tabular}
\caption{5-gradient update AUC results for Ego-Aminer for training edge splits}
\end{center}
\vskip -0.1in
\end{table*}

\cut{
\begin{table*}[ht]
%\label{ml-table}
\begin{center}
\begin{tabular}{lccccccccr}
\toprule
PPI 
  &  $10\%$ & $20\%$ &  $30\%$ & \textbf{$40\%$-high p-val}& $50\%$ & $60\%$ & $70\%$\\
\midrule
Log Degree \\ Difference & NaN& 0.586/0.807& 0.493/0.699 & -0.063/-0.045 & -0.364/-0.368 &-0.638/-0.688 & -0.796/-0.851\\
Avg-Emb &NaN & -0.508/-0.891& -0.469/-0.847 & 0.008/-0.018 & 0.298/0.186 & 0.555/0.522& 0.860/0.660 \\
WL-kernel &NaN & -0.298/-0.284& -0.299/-0.276& -0.155/-0.132 & \textbf{high p-val}& \textbf{high p-val} & 0.342/0.245\\

\bottomrule
\end{tabular}
\end{center}
\vskip -0.1in
\end{table*}
}

\cut{
\begin{table*}[ht]
%\label{ml-table}
\begin{center}
\begin{tabular}{lccccccccr}
\toprule
Aminer-20k\\
Convergence  &  $10\%$ & $20\%$ &  $30\%$ & $40\%$  & $50\%$ & $60\%$ & $70\%$\\
\midrule
\textbf{Meta-Graph} & 0.681/0.656 & 0.738/0.719 & 0.790/0.775 & 0.825/0.813 & 0.847/0.834 & 0.863/0.855& 0.856/0.843\\
MG-Random &  0.500/0.500 & 0.500/0.500 & 0.500/0.500 & 0.500/0.500 & 0.500/0.500 & 0.500/0.500 & 0.500/0.500\\
\textbf{MAML}    & 0.690/0.660 & 0.751/0.729 & 0.793/0.757 & 0.831/0.805 & 0.828/0.804& 0.861/0.841 &0.857/0.840\\
Random    &  0.500/0.500 & 0.500/0.500 & 0.500/0.500 & 0.500/0.500 & 0.500/0.500 & 0.500/0.500 & 0.500/0.500\\
\textbf{No Finetune} & 0.675/0.657& 0.788/0.776&0.834/0.831& 0.859/0.853 &0.890/0.891 &0.877/0.867 & 0.835/0.833\\
\textbf{Finetune} & 0.812/0.800 & 0.879/0.873 & 0.890/0.881 & 0.905/0.902 & 0.912/0.910 & 0.931/0.929 & 0.942/0.942\\
Adamic    & 0.514/0.514 & 0.553/0.553 & 0.606/0.607 & 0.664/0.664 & 0.720/0.721 &0.772/0.773 &0.814/0.815\\
MAML-MLP    &0.701/0.696 &0.711/0.706 &0.703/0.6983 &0.714/0.707 &0.710/0.704 &0.712/0.703 & 0.711/0.705 \\
Deepwalk    & 0.588/0.632 & 0.714/0.781 & 0.829/0.873& 0.898/0.921& 0.931/0.944 & 0.949/0.956 & 0.960/0.964 \\

\bottomrule
\end{tabular}
\end{center}
\vskip -0.1in
\end{table*}

\begin{table*}[ht]
%\label{ml-table}
\begin{center}
\begin{tabular}{lccccccccr}
\toprule
Aminer-20k \\
5 updates  &  $10\%$ & $20\%$ &  $30\%$ & $40\%$  & $50\%$ & $60\%$ & $70\%$\\
\midrule
\textbf{Meta-Graph} & \\
\textbf{MAML} & \\
\textbf{No Finetune} & 0.500/0.500 & 0.500/0.500 & 0.500/0.500 & 0.500/0.500 & 0.500/0.500 &  0.500/0.500 & 0.500/0.500\\
\textbf{Finetune} & 0.702/0.684 & 0.821/0.811 & 0.847/0.838& 0.853/0.850 & 0.884 & 0.858/0.854 & 0.886/0.884\\
MAML-MLP &  0.601/0.595 & 0.597/0.59 & 0.601/0.592& 0.613/0.612 & 0.878 & 0.614/0.608 & 0.630/0.623\\

\bottomrule
\end{tabular}
\end{center}
\vskip -0.1in
\end{table*}
}

\end{document}